%% file: iclr2026_conference.tex
\documentclass{article} 
\usepackage{iclr2026_conference,times}

\input{math_commands.tex}

\definecolor{softblue}{rgb}{0.21,0.49,0.74}
\usepackage[breaklinks,colorlinks,allcolors=softblue]{hyperref}
\usepackage{url}
\usepackage{graphicx}
\usepackage{algorithm}
\usepackage{algorithmic}
\usepackage{cuted}
\usepackage{caption}
\usepackage{booktabs}
\usepackage{multirow}
\usepackage{amsmath}
\usepackage{amssymb}
\usepackage{mathtools}
\usepackage{subcaption}
\usepackage{wrapfig}

\usepackage{xcolor} 
\definecolor{commentblue}{rgb}{0.1, 0.5, 0.7}
\newcommand{\algcomment}[1]{\textcolor{commentblue}{\# #1}}

\usepackage{xcolor}
\definecolor{mypurple}{RGB}{128, 0, 128}

\title{\textcolor{mypurple}{$\mathbf{S}$}tochastic \textcolor{mypurple}{$\mathbf{S}$}elf\textcolor{mypurple}{-$\mathbf{G}$uidance} for Training-Free Enhancement of Diffusion Models}

\iclrfinalcopy 
\begin{document}

\maketitle
\vspace{-6em}

\begin{center}
    \textbf{Chubin Chen\textsuperscript{1,*}} \quad
    \textbf{Jiashu Zhu\textsuperscript{2}} \quad
    \textbf{Xiaokun Feng\textsuperscript{3}} \quad
    \textbf{Nisha Huang\textsuperscript{1}} \quad
    \textbf{Chen Zhu\textsuperscript{2}} \\
    \textbf{Meiqi Wu\textsuperscript{3}} \quad
    \textbf{Fangyuan Mao\textsuperscript{3}} \quad
    \textbf{Jiahong Wu\textsuperscript{2,‡}} \quad
    \textbf{Xiangxiang Chu\textsuperscript{2}} \quad
    \textbf{Xiu Li\textsuperscript{1,†}} \vspace{0.5em} \\
    \textsuperscript{1}Tsinghua University \quad
    \textsuperscript{2}AMAP, Alibaba Group \quad
    \textsuperscript{3}CASIA \vspace{0.5em} \\
    \textit{Project Page}:
    \href{https://s2guidance.github.io/}
    {\texttt{\textcolor{mypurple}
    {\textit{https://s2guidance.github.io/}}}}
\end{center}

\renewcommand{\thefootnote}{\fnsymbol{footnote}} 
\footnotetext[1]{Work done during the internship at AMAP, Alibaba Group.}
\footnotetext[2]{Corresponding author.} 
\footnotetext[3]{Project lead.} 



\begin{figure}[h]
\centering
\vspace{-10pt}
    \includegraphics[width=1.0\linewidth]{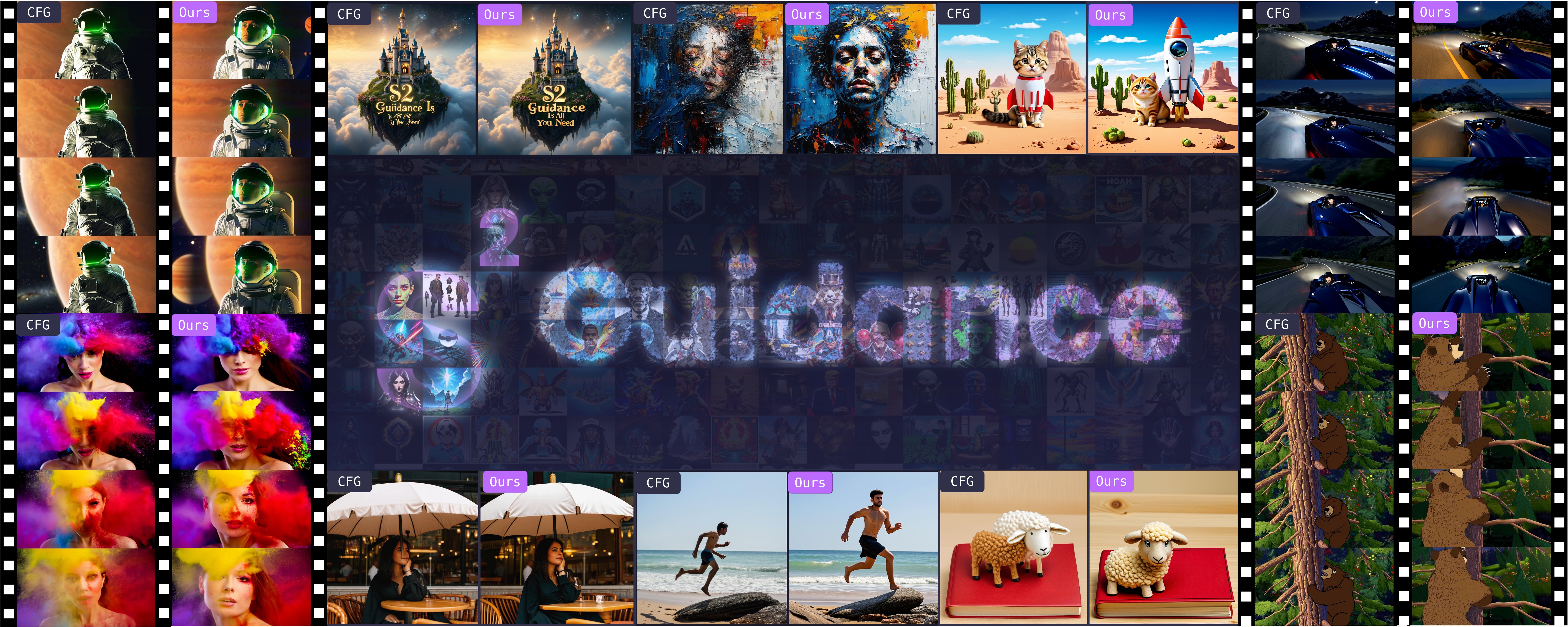}
\captionof{figure}{\textbf{Visual results of \ours versus CFG.}
    Our proposed method \ours significantly elevates the quality and coherence of both T2I and T2V generation. 
    \textbf{Observe (in examples surrounding the center):} 
    Our method produces generations with \textbf{superior temporal dynamics}, including more pronounced motion (bear) and dynamic camera angles that convey speed (car). It renders \textbf{finer details}, such as the astronaut's transparent helmet and rich facial details, and creates images with \textbf{fewer artifacts} (runner, woman with umbrella), \textbf{richer artistic detail} (abstract portrait, castle, colored powder exploding), and \textbf{improved object coherence} (cat and rocket, sheep). See Appendix \ref{app:figure1_prompts} for our prompts.
}
\label{fig:first_pic}
\vspace{-10pt}
\end{figure}

\input{sec/0_abs}
\input{sec/1_intro}

\input{sec/2_related}
\input{sec/3_method}

\input{sec/4_exp}

\input{sec/5_conclusion}

\bibliography{iclr2026_conference}
\bibliographystyle{iclr2026_conference}

\appendix
\section{Appendix}
\input{sec/6_appendix}

\end{document}

%% file: math_commands.tex
\usepackage{amsmath,amsfonts,bm}

\usepackage{xspace}
\newcommand{\ours}{{$S^2$-Guidance}\xspace}
\newcommand{\eg}{\textit{e}.\textit{g}.}

\def\eqref#1{equation~\ref{#1}}

\def\1{\bm{1}}

\DeclareMathAlphabet{\mathsfit}{\encodingdefault}{\sfdefault}{m}{sl}
\SetMathAlphabet{\mathsfit}{bold}{\encodingdefault}{\sfdefault}{bx}{n}



%% file: sec/0_abs.tex
\begin{abstract}

Classifier-free Guidance (CFG) is a widely used technique for improving conditional generation in diffusion models.
However, our empirical analysis of both Gaussian mixture data and real-world image data distributions reveals a discrepancy between the suboptimal results produced by CFG and the ground truth. The model's excessive reliance on these suboptimal predictions often leads to low fidelity and semantic incoherence. 
To address this issue, we first empirically demonstrate that the model's suboptimal predictions can be effectively rectified using sub-networks of the model itself, without requiring additional training or the integration of external modules. 
Building on this insight, we propose \ours ($S$tochastic $S$elf-Guidance), a novel method that leverages stochastic block-dropping during the denoising process to activate sub-networks for self-guidance. This approach effectively steers the sampling trajectory towards high-quality regions. 
Comprehensive experiments, including on class-conditional ImageNet generation and across multiple benchmarks for text-to-image and text-to-video generation, demonstrate the superiority of \ours. Both qualitative and quantitative results show that \ours consistently surpasses CFG and other advanced guidance strategies. Our code will be released.
\end{abstract}

%% file: sec/1_intro.tex
\section{Introduction}

Diffusion models \citep{song2020denoising,ho2020denoising} have enabled rapid advances in high-quality text-to-image  \citep{rombach2022high,podell2023sdxl,flux,wu2025qwenimagetechnicalreport} and text-to-video \citep{wan2025wan,kong2024hunyuanvideo} generation.
A key driver of this success is the advent of conditional guidance techniques, which steer the generation process to enhance adherence to given conditions. However, naively applying the conditioning signal often proves insufficient \citep{diffusionbeatsgans}.
Classifier-free Guidance (CFG)~\citep{cfg} has become the mainstream approach for improving conditional generation.
It employs a Bayesian implicit classifier to prioritize conditional probability, enhancing adherence to conditions and image quality.
However, despite its effectiveness, it often results in semantic incoherence and a loss of fine details, as shown in Figure~\ref{fig:first_pic}.

Recent studies~\citep{cfg++,apg,cfgzero,intervalcfg,adg} have further explored methods to improve guidance.
Although these methods improve quality to some extent, they primarily address specific issues while leaving the underlying mechanisms of CFG unexplored.
A representative work that begins to explore this issue is Autoguidance~\citep{Autoguidance}, which identifies deficiencies in the model's training objective and proposes using a weak model for guidance. Subsequent works~\citep{sag,pag,spg,seg} propose modifying specific attention regions to mimic a weak model for various tasks \citep{attn1,attn2}.
However, these methods either require training to acquire the weak model or rely on empirical, task-specific modifications to the network, which in turn demand meticulous hyperparameter tuning.

To address this, we first analyze the suboptimal results produced by CFG and the underlying mechanisms of weak-model guidance. 
Specifically, our analysis begins with a toy example on Gaussian mixture modeling, where a closed-form solution allows for precise evaluation against the ground truth~\citep{gauss1,gauss2}, and is subsequently validated on real-world image data.
Furthermore, we observe that applying stochastic block-dropping during the model's forward process produces results highly similar to the weak model used in Autoguidance. 
Building on this discovery, we propose \ours, a simple yet effective approach to address the suboptimal predictions of CFG and guide sampling towards higher quality and fidelity.
Unlike prior methods that rely on externally trained or manually tuned weak models, \ours leverages the model's own intrinsic structure in a training-free manner, effectively steering the denoising trajectory away from failure modes to enhance the performance of conditional diffusion models.

Our contributions are summarized as follows: 

\begin{wrapfigure}{12}{0.65\textwidth}
    \centering
    \vspace{-10pt}
    \includegraphics[width=0.65\textwidth]{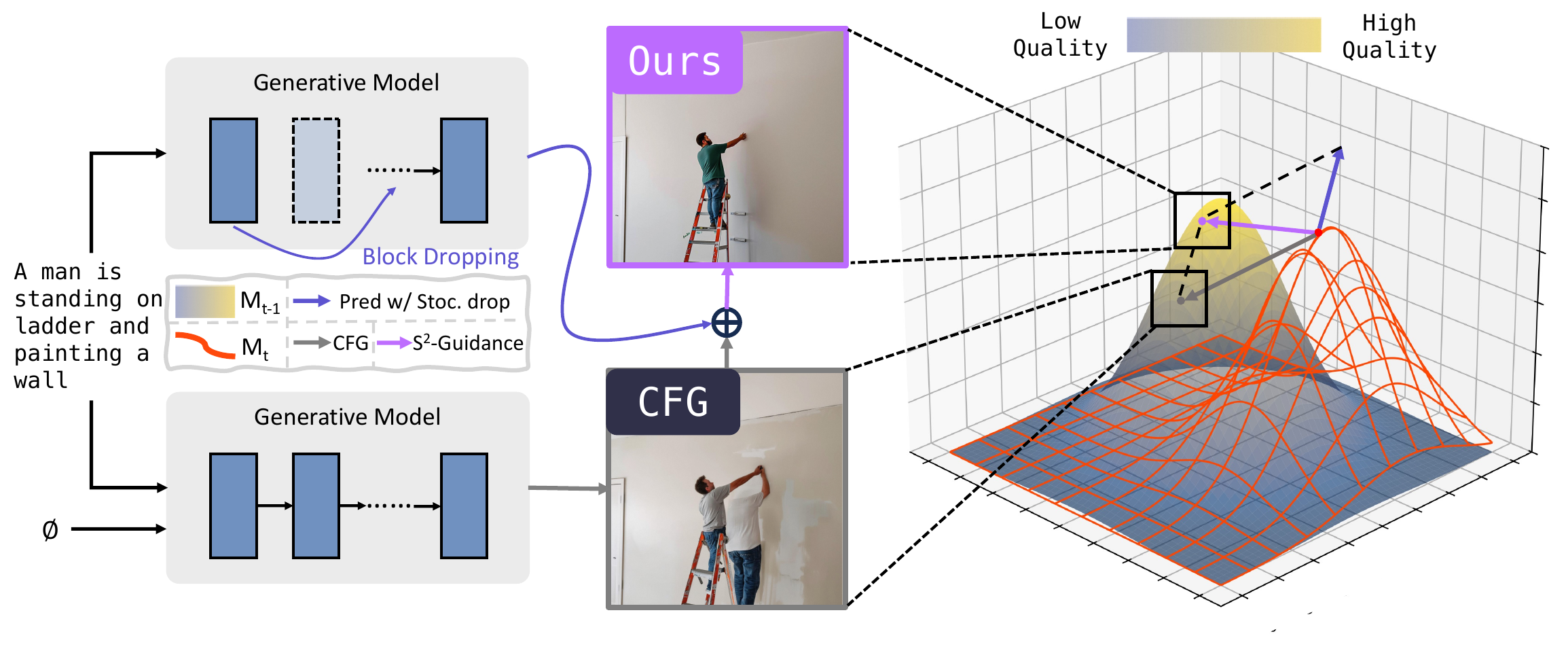}
    \captionof{figure}{
        \textbf{An illustration of our guidance mechanism on the generation quality manifold.} 
    Unlike suboptimal CFG guidance (gray), \ours derives a corrective signal (blue) via stochastic block-dropping, steering the generation update (purple) toward the optimal quality peak (yellow).}
    \label{fig:theory}
\end{wrapfigure}

\textbf{(i)} We first analyze the guidance behavior of CFG and the underlying mechanisms of weak-model guidance through a series of toy examples. These examples allow us to visually analyze the suboptimal results of CFG. Empirical observations reveal that the sampling trajectory can be effectively rectified by the model's own sub-networks, which exhibit guidance behavior similar to that of a weak model.

\textbf{(ii)} We propose \ours, a novel method that leverages stochastic block-dropping during the forward process to activate sub-networks for self-guidance, thereby bypassing the need to construct weak models through additional training or a trial-and-error manual selection process. 
Furthermore, we demonstrate that in the iterative denoising process, a single block-dropping per timestep is sufficient to steer the sampling trajectory towards high-quality regions. This approach achieves strong performance while substantially reducing computational costs compared to the naive variant.

\textbf{(iii)} Our method can be seamlessly adapted to various diffusion models. 
Comprehensive experiments—on class-conditional ImageNet generation and across multiple benchmarks for text-to-image and text-to-video tasks—establish the superiority of \ours. Both qualitative and quantitative results confirm that \ours consistently surpasses not only CFG but also other advanced guidance strategies.

%% file: sec/2_related.tex
\begin{figure*}[t!]
\centering
\includegraphics[width=1.0\textwidth]{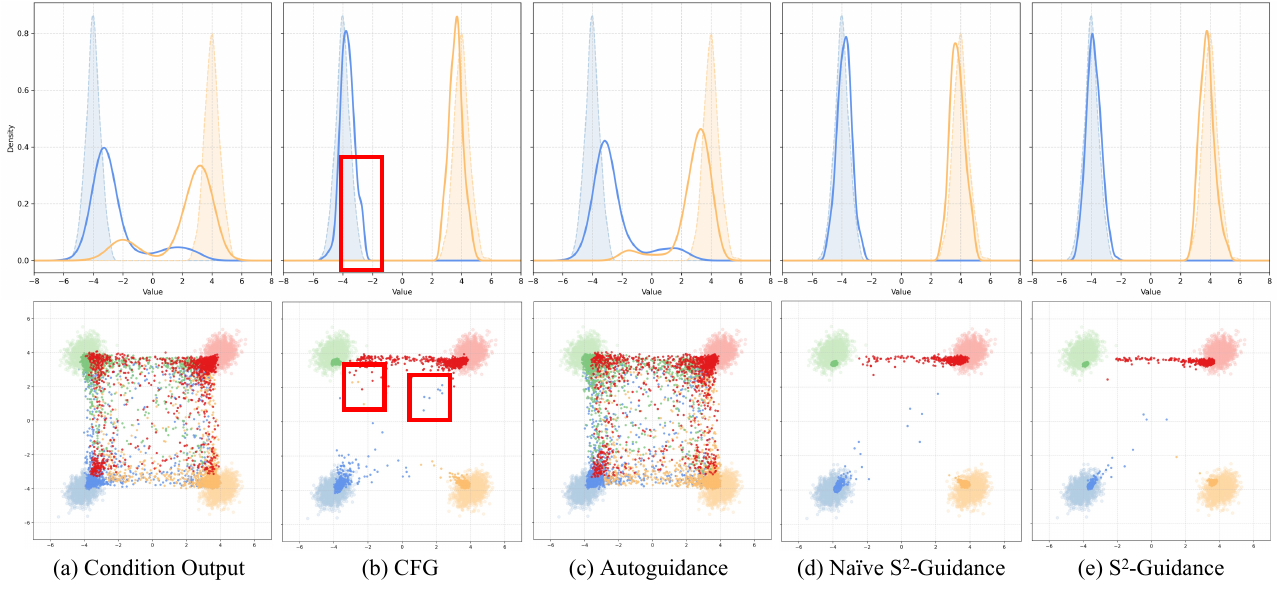}
\caption{
    \textbf{\ours~successfully balances guidance strength and distribution fidelity.}
    Comparison on 1D (top) and 2D (bottom) toy examples. 
    Unlike CFG, which distorts the sample distribution (see red boxes), or other methods that fail to separate modes, \ours~accurately captures both the location and shape of the ground truth distributions (semi-transparent).
}
\label{fig:1d_toy_distribution}
\vspace{-10pt}
\end{figure*}

\section{Background}

\paragraph{Diffusion Models.}
Diffusion models \citep{croitoru2023diffusion,peebles2023scalable,sd3, chu2025usp} are a class of powerful generative models that learn to reverse a predefined forward process, which gradually perturbs data $x_0$ into Gaussian noise $x_T$. 
The reverse process is typically governed by a time-reversed stochastic differential equation (SDE) \citep{sde}, which relies on accurately estimating a score function $\nabla_{x_t} \log p_t(x_t)$ using a neural network $D_\theta$.
Flow-based models \citep{flow1, flow2,flow3} can also be viewed as a special class of diffusion models, as they both aim to learn a continuous transformation between a simple prior distribution and the complex data distribution \citep{flow5}.
In practical applications \citep{huang2023t2i, zhu2024instantswap, huang2024diffstyler, omni}, generation is often conditioned on signals $c$ (\eg, text prompts), shifting the objective to modeling the conditional score $\nabla_{x_t} \log p_t(x_t|c)$.

\paragraph{Classifier-free Guidance (CFG).}
CFG \citep{cfg} has become the cornerstone for controllable generation \citep{huang2023region, wang2024taming, fang2025integrating, he2025diffusion, ma2024followpose, ma2025followyourclick} by offering a simple yet effective mechanism to enhance conditioning. 
Instead of only using the conditional prediction $D_\theta(x_t|c)$, CFG forms a guided score by extrapolating from an unconditional one $D_\theta(x_t|\phi)$:
\begin{align} \label{eq:cfg}
    \tilde{D}^{\lambda}_\theta(x_t|c) = D_\theta(x_t|\phi) + \lambda \left( D_\theta(x_t|c) - D_\theta(x_t | \phi) \right),
\end{align}
where $\lambda$ is the guidance scale.
However, despite its effectiveness, this approach suffers from notable drawbacks \citep{apg, sag, Autoguidance}, including semantic inconsistencies and a significant loss of fine-grained details, as illustrated in Figure~\ref{fig:first_pic}.


\paragraph{Weak-model Guidance.}
A promising direction to improve CFG is to leverage an auxiliary ``weak" model to refine the guidance signal.
For instance, Autoguidance \citep{Autoguidance} employs a separately trained, degraded version of the full model, but such models are often infeasible to obtain for large-scale pretrained models.
To circumvent this, recent works simulate a weak model by modifying the model's architecture or perturbing its internal states. 
For instance, some studies rely on heuristic perturbations like attention-guided blurring of predicted samples \citep{sag,pag}; SEG \citep{seg} later proposes an alternative from an energy-based perspective; and other works develop strategies for specific tasks \citep{spg, stg}
However, these perturbation techniques often rely on task-specific, hand-crafted architectural modifications,
which limits their generalizability. 
In contrast, as shown in Figure \ref{fig:theory}, our \ours introduces a novel and flexible approach. We guide the sampling process by dynamically activating sub-networks via stochastic block dropping, thereby avoiding the need to construct a weak model through auxiliary training or manually designed perturbation schemes.

%% file: sec/3_method.tex
\section{Methodology}

\begin{figure*}[t!]
\centering
\includegraphics[width=1.0\textwidth]{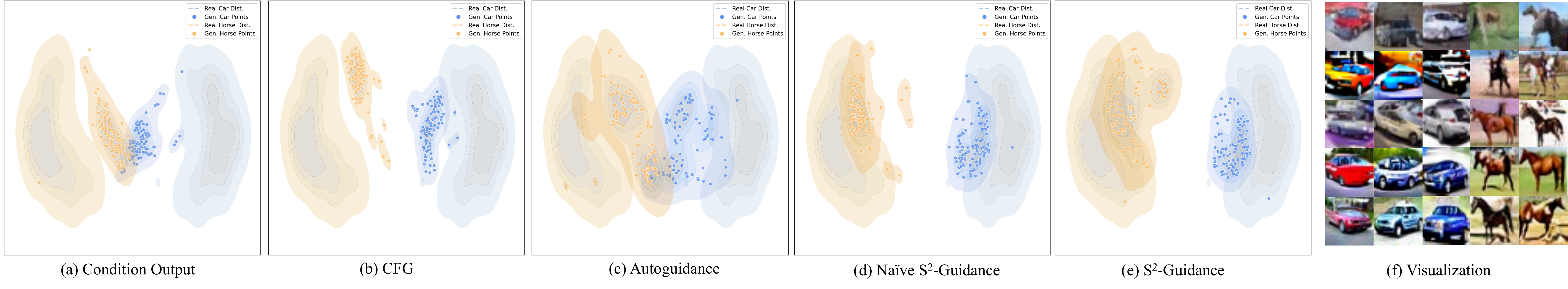}
\caption{
    \textbf{\ours{} avoids the distributional collapse of CFG on CIFAR-10.} 
    t-SNE shows generated features (points) vs. real data (contours). CFG (b) exhibits severe collapse, whereas \ours{} (e) preserves the distribution's structure while ensuring class separation. See (f) for qualitative examples.
}
\label{fig:cifar_dist}
\vspace{-10pt}
\end{figure*}

\subsection{Visualizing and Revisiting Weak-model Guidance}
We begin by visualizing the suboptimal outcomes of CFG using Gaussian mixture data \citep{cfg}, a toy example with closed-form solutions. This allows us to systematically observe the discrepancies between predictions and ground truth. 
Building on the analysis of how weak-model guidance \citep{Autoguidance} improves results, we identify its limitations and propose incorporating stochastic sub-networks into the CFG framework, providing a novel approach to enhance model performance.

CFG improves conditional generation by implicitly amplifying the conditional probability density, raising it to a power greater than one \citep{bradley2024classifier}. 
Figure \ref{fig:1d_toy_distribution} illustrates a 1D toy example (top) aimed at learning a Gaussian Mixture distribution with modes at $-4$ and $4$. While CFG significantly improves the baseline conditional output, it also introduces a notable drawback: as highlighted by the red box, the mode of the generated distribution is slightly shifted from the ground truth.
A similar shift occurs in a 2D toy example (bottom), where samples are scattered into unintended regions. These findings suggest that, although CFG enhances sample quality, its distributional fidelity remains suboptimal.
Autoguidance, as a representative of weak-model guidance~\citep{Autoguidance,sag,seg,pag}, is designed to guide the model toward well-learned, high-probability regions by leveraging a weak model. 
As shown in Figure~\ref{fig:1d_toy_distribution} (middle), AutoGuidance improves the peak near -4 but remains limited. Its improvement stems from the construction of a weak model, with the extent of enhancement depending on the weak model's effectiveness.
Such models are typically created by reducing model capacity or training epochs. 

However, this approach faces practical limitations that restrict its broader applicability.
First, relying on externally designed weak models poses scalability challenges, as obtaining a reduced version trained for fewer epochs alongside a large-scale pretrained model is often impractical.
Second, as highlighted by \citep{Autoguidance}, selecting an appropriate weak model is constrained by various factors. Once chosen, the weak model affects the entire denoising process, limiting the flexibility of guidance. 
A poorly designed weak model fails to effectively prevent low-quality outputs \citep{seg}, as shown in Figure~\ref{fig:1d_toy_distribution} (c), where guided outputs still deviate notably from the target distribution.

This raises an important question: \textit{Can we eliminate the reliance on externally prescribed weak models while still identifying error-prone regions?}
Prior works~\citep{redundant1,redudant2_stableflow,redundant3} have shown that mainstream generative architectures, such as DiT~\citep{peebles2023scalable,chu2024visionllama}, exhibit significant redundancy, as outputs across different transformer blocks often show high similarity~\citep{chen2024revealing}. 
Inspired by this, we hypothesize that sub-networks within such architectures can function as weak models, capturing outputs similar to the full model but with more pronounced errors. By leveraging these sub-network predictions, we aim to refine existing CFG, effectively steering the model away from suboptimal outputs.
The following subsections present a detailed description of our approach along with its empirical validation.

\subsection{Naive \ours}
Building on the preceding observation, our key insight is that \textit{we can leverage the model’s own sub-networks to intrinsically steer the denoising trajectory away from potential failure modes, thereby refining the suboptimal results of CFG}.

As revealed in Autoguidance~\citep{Autoguidance}, problems in generative models depend on various factors (e.g., network architecture, dataset properties, etc.), making it difficult to pinpoint which components play a decisive role.
Therefore, it is challenging to \textit{a priori} define an optimal sub-network that best captures low-quality regions. Motivated by~\citep{dropout}, a naive solution is to leverage as many diverse stochastic sub-networks as possible to construct multiple weak models. These weak models then guide the main model away from low-quality regions during each forward pass by steering it away from their outputs.
We refer to this approach as Naive $S$tochastic $S$ub-network Guidance (Naive \ours).
Intuitively, this can be understood as applying stochastic ``dropout" to different blocks, constructing various sub-networks that capture diverse low-probability regions.


Specifically, for a given binary mask \( \mathbf{m} \), sampled via stochastic block-dropping from the induced distribution \( p(\mathbf{m}) \), the weak model's prediction is defined as:
\begin{align}
\hat{D}_\theta(x_t \mid c, \mathbf{m}) = D_\theta(x_t \mid c; \boldsymbol{\theta} \odot \mathbf{m}),
\end{align}
where \( \mathbf{m} \) determines which blocks of the network parameters \( \boldsymbol{\theta} \) are activated, forming a latent sub-network during each forward pass.
Naive \ours is then expressed as:
\begin{align}
    \tilde{D}_\theta^\lambda(x_t \mid c)
    = &\; D_\theta(x_t \mid \phi)
    +  \lambda \big( D_\theta(x_t \mid c) - D_\theta(x_t \mid \phi) \big) \notag \\
    & - \frac{\omega}{N} \sum_{i=1}^N
    \big(
      \hat{D}_\theta(x_t \mid c, \mathbf{m}_i)
      - D_\theta(x_t \mid c)
    \big),
\end{align}
where \( \mathbf{m}_i \sim p(\mathbf{m}) \) is the binary mask for the \( i \)-th stochastic sub-network, \( \omega \) controls the strength of the self-guidance, referred to as the \( S^2 \) Scale. \( \hat{D}_\theta(x_t \mid c, \mathbf{m}_i) \) represents the prediction from the \( i \)-th sampled sub-network, and the self-guidance signal is defined as its deviation from the full-model prediction. \( N \) denotes the total number of latent sub-networks sampled during each forward pass.

For the sampling distribution \( p(\mathbf{m}) \), a crucial consideration is to ensure its effectiveness and generalizability across different models. Our approach is predicated on the principle of identifying and preserving the model's structurally critical components. Based on empirical analysis, we exclude these key blocks from the dropping process and then sample a proportion of the remaining blocks to be dropped. 
Detailed ablations regarding the stochastic sampling design and its stability are provided in Section~\ref{sec:exp}.

To validate our hypothesis, we conduct experiments on toy examples with 1D and 2D Gaussian mixture data, as well as on real-world datasets (see Appendix \ref{app:toy} for more details). As shown in Figure~\ref{fig:1d_toy_distribution} (d), compared to the original CFG, our Naive \ours not only leads to predictions that better fit the target distribution but also mitigates the drift phenomenon, thereby improving fidelity. This demonstrates that our method effectively refines the suboptimal results of CFG. 
Furthermore, compared to Autoguidance, \ours eliminates the need for explicitly constructing weak models. By adopting this simple yet effective approach, it avoids generating results that lie in intermediate regions, thereby reducing mode confusion.
These results provide strong empirical evidence that leveraging Naive \ours can significantly enhance both the quality and robustness of conditional generation.

\begin{table*}[t!]
\centering
\resizebox{1.0\textwidth}{!}{
\setlength{\tabcolsep}{4pt}

\begin{tabular}{l|l|ccccc|ccc|cc} 
\toprule
\multicolumn{1}{c|}{\multirow{2}{*}{Model}} & \multicolumn{1}{c|}{\multirow{2}{*}{Method}} & \multicolumn{5}{c|}{HPSv2.1 (\%) $\uparrow$} & \multicolumn{3}{c|}{T2I-CompBench (\%) $\uparrow$} & \multicolumn{2}{c}{Qalign $\uparrow$} \\
\cmidrule(lr){3-7} \cmidrule(lr){8-10} \cmidrule(lr){11-12}
& & Anime & Concept & Paint. & Photo & Avg. & Color & Shape & Texture & HPSv2.1 & T2I-Comp. \\ \midrule

\multirow{5}{*}{SD3} 
& CFG       & 31.55 & 30.87 & 31.22 & 28.27 & 30.48 & 53.61 & 51.20 & 52.45 & \underline{4.66} & \underline{4.74} \\
& CFG++     & 31.57 & 30.76 & 30.96 & 27.54 & 30.21 & 46.39 & 47.18 & 46.33 & \textbf{4.68} & 4.73 \\
& APG       & 30.77 & 30.18 & 30.53 & 27.12 & 29.65 & 45.28 & 46.27 & 46.84 & \textbf{4.68} & 4.73 \\
& CFG-Zero  & \underline{31.99} & \underline{31.17} & \underline{31.42} & 28.54 & \underline{30.78} & 52.70 & 52.84 & 53.37 & \underline{4.66} & \textbf{4.77} \\
& SEG       & 31.20 & 30.56 & 31.07 & \underline{28.74} & 30.39 & \underline{58.20} & \underline{57.68} & \textbf{57.17} & 4.33 & 4.45 \\
& \textbf{Ours} & \textbf{32.14} & \textbf{31.32} & \textbf{31.70} & \textbf{29.19} & \textbf{31.09} & \textbf{59.63} & \textbf{58.71} & \underline{56.77} & 4.65 & \underline{4.74} \\
\midrule
\multirow{5}{*}{SD3.5} 
& CFG       & 32.34 & 31.51 & 31.50 & 27.93 & 30.82 & 51.29 & 47.71 & 47.39 & 4.63 & 4.66 \\
& CFG++     & 31.99 & 31.02 & 31.36 & 27.32 & 30.42 & 38.05 & 37.52 & 34.87 & 4.65 & 4.58 \\
& APG       & 31.43 & 30.74 & 31.12 & 27.07 & 30.09 & 35.67 & 37.86 & 35.67 & \underline{4.68} & 4.65 \\
& CFG-Zero  & \underline{32.77} & \underline{31.91} & \underline{31.95} & 28.27 & \underline{31.23} & 52.01 & 46.99 & 48.36 & 4.66 & \underline{4.70} \\
& SEG       & 31.77 & 31.30 & 31.40 & \underline{28.34} & 30.71 & \textbf{57.59} & \textbf{55.52} & \textbf{54.03} & 4.41 & 4.45 \\
& \textbf{Ours} & \textbf{32.89} & \textbf{32.15} & \textbf{32.28} & \textbf{28.94} & \textbf{31.56} & \underline{57.57} & \underline{51.23} & \underline{50.13} & \textbf{4.70} & \textbf{4.74} \\
\bottomrule
\end{tabular}
}
\caption{\textbf{Quantitative comparison in T2I generation.}
    Our method establishes a new state-of-the-art, demonstrating significant improvements even on highly competitive benchmarks. 
    On \textbf{HPSv2.1}, a benchmark where score margins are typically narrow, \ours{} consistently outperforms all baselines across every individual dimension.
    This lead is even more pronounced on \textbf{T2I-CompBench}, where our approach shows substantial gains in compositional attributes like Color and Shape.
    Notably, \ours also achieves the highest or near-highest aesthetic scores (\textbf{Qalign}) on both benchmarks, demonstrating its superior performance in visual quality.
    Higher scores ($\uparrow$) are better. Best results are in \textbf{bold}; second-best are \underline{underlined}.
}
\label{tab:hpsv2_t2i_full_fixed}
\end{table*}

\subsection{\ours Is Sufficient}
However, Naive \ours incurs significant computational overhead, which severely limits its practicality.
In the process of constructing sub-networks, we find that constraining stochastic block-dropping within a specific range allows sub-networks, even those generated by dropping at different blocks, to consistently guide the model toward the ideal distribution (Figure \ref{fig:toy_trajectories}).

Therefore, we propose a simplified approach: performing a single stochastic block-dropping operation at each timestep for self-guidance. We refer to this approach as \ours, which achieves highly competitive results.
At timestep \( t \), \ours is expressed as:
\begin{align}
    \tilde{D}_\theta^\lambda(x_t \mid c) = & \; D_\theta(x_t \mid \phi)
    + \lambda \big( D_\theta(x_t \mid c) - D_\theta(x_t \mid \phi) \big) \notag \\
    & - \omega \big(  \hat{D}_\theta(x_t \mid c, \mathbf{m}_t) 
    - D_\theta(x_t \mid c)
    \big ).
\end{align}
The overall algorithm is summarized in Algorithm \ref{alg:one}.

\begin{wrapfigure}[14]{r}{0.6\textwidth} 

  \vspace{-22pt} 

  \begin{minipage}{\linewidth} 

    \begin{algorithm}[H] 

    \caption{\ours}

    \label{alg:one}

    \begin{algorithmic}[1]

        \REQUIRE Trained denoiser $D_\theta$, initial noise $x_T$, guidance scale $\lambda$, $S^2$ scale $\omega$, number of timesteps $T$.

        \FOR{$t = T, \dots, 1$}

            \STATE $m_t \gets \text{GenerateStochasticMask}()$ \algcomment{Generate mask}

            \STATE $D_\text{uncond} \gets D_\theta(x_t, \phi, t)$

            \STATE $D_\text{cond} \gets D_\theta(x_t, c, t)$

            \STATE $\hat{D}_\text{s} \gets D_\theta(x_t, c, t, m_t)$ \algcomment{Sub-network prediction}

            \STATE $\tilde{D} \gets D_\text{uncond} + \lambda(D_\text{cond} - D_\text{uncond}) - \omega (\hat{D}_\text{s} - D_\text{cond})$

            \STATE $x_{t-1} \gets \text{SchedulerStep}(\tilde{D}, x_t, t)$

        \ENDFOR

        \RETURN $x_0$

    \end{algorithmic}

    \end{algorithm}

  \end{minipage}

\end{wrapfigure}

We empirically validate the proposed \ours on toy examples with 1D and 2D Gaussian mixture data, as well as on real-world datasets. 
As shown in Figure~\ref{fig:1d_toy_distribution} (e), \ours performs comparably to Naive \ours. On both 1D and 2D Gaussian mixture distributions, it produces results that closely align with the ideal distribution, while exhibiting efficiency without significant degradation.
Moreover, as illustrated in Figure~\ref{fig:cifar_dist} (e, f), \ours achieves highly competitive performance on real-world datasets, highlighting its practical effectiveness.
To further analyze the stochastic block-dropping strategy, we conduct a detailed experimental study in Section~\ref{exp:ab_study}. 
Our empirical analysis reveals that, when the block drop ratio is maintained around 10\% of the network's blocks, the resulting sub-networks consistently enable the model to achieve better performance.
This strategy proves effective across mainstream DiT \citep{peebles2023scalable} architectures, leveraging the redundancy in the outputs to dynamically construct diverse stochastic sub-networks. Unlike explicitly constructed weak models, which once selected affect the entire denoising process, stochastic block-dropping enables the creation of sub-networks independently at different timesteps. This dynamic diversity introduces self-guidance throughout the diffusion process, allowing predictions to evolve iteratively and steering the outputs toward higher-quality results.

%% file: sec/4_exp.tex
\newcommand{\rot}[1]{\begin{turn}{60}\textbf{#1}\end{turn}}
\section{Experiments}
\label{sec:exp}

\begin{figure*}[t]
\centering
\includegraphics[width=1.0\textwidth]{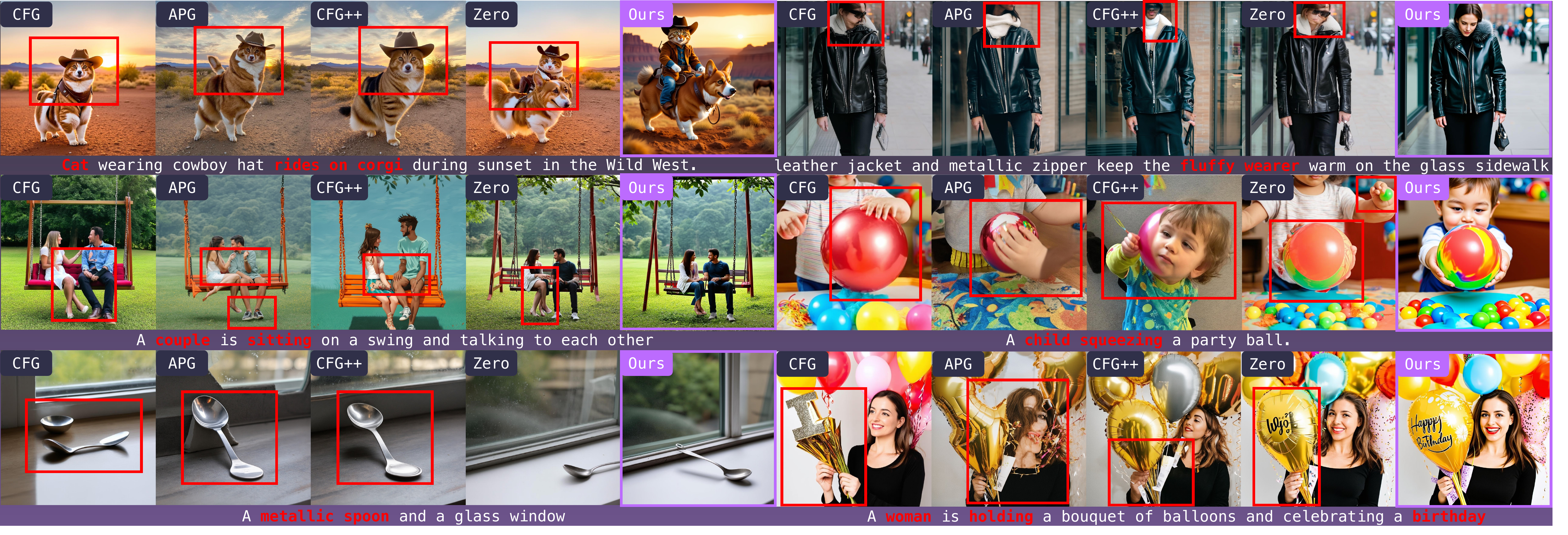}
\caption{
        \textbf{\ours consistently generates superior images in both aesthetic quality and prompt coherence.}
    While existing guidance methods often produce artifacts, distorted objects, or fail to follow complex prompts (see red boxes), our approach yields clean, coherent, and visually pleasing results without such flaws.
    }
\label{fig:t2i_qualify}
\end{figure*}

\begin{figure*}[t]
\centering
\includegraphics[width=1.0\textwidth]{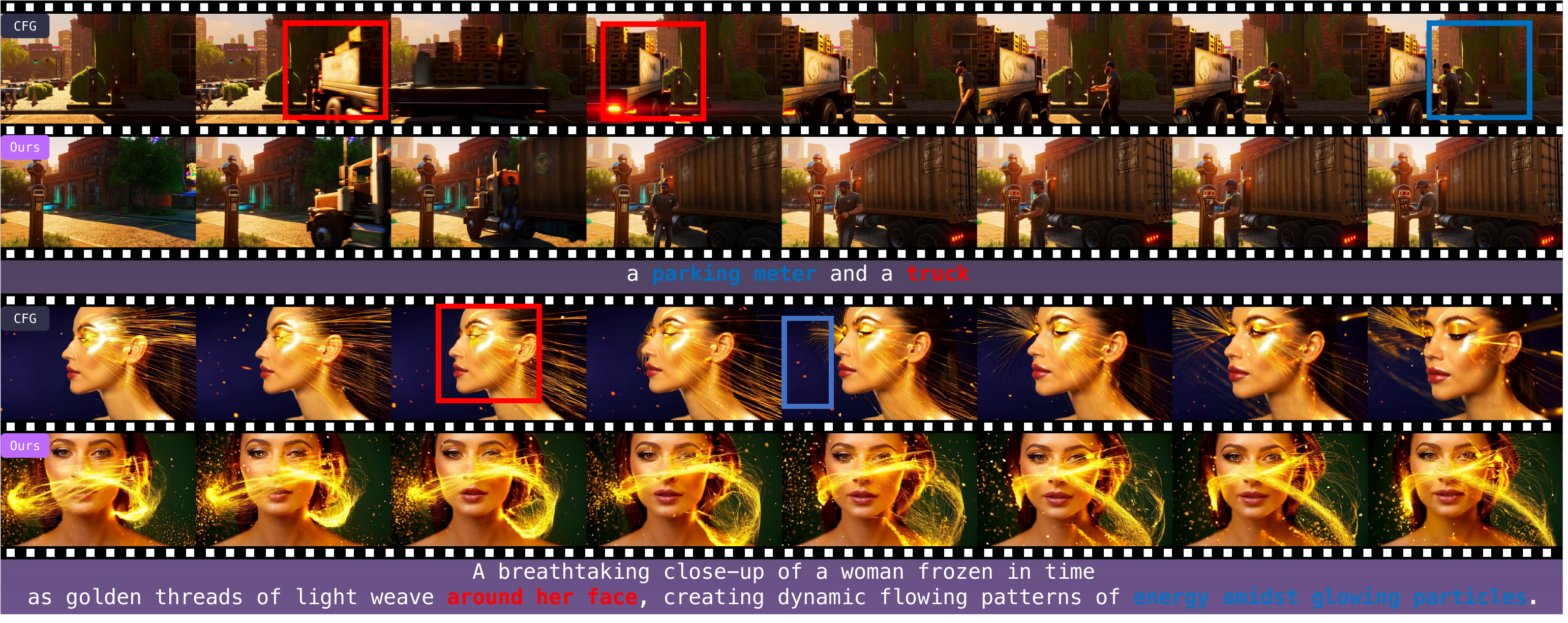}
\caption{
        \textbf{\ours{} generates temporally coherent and physically plausible videos, overcoming key failures of CFG.}
    \textbf{Top Row:} CFG struggles with plausible motion, depicting a truck that unnaturally slides sideways instead of driving forward (red boxes). Our method renders a stable and realistic scene.
    \textbf{Bottom Row:} CFG fails to capture the full prompt, as the light does not weave “around her face” (red box) and lacks “glowing particles” (blue box). \ours{} faithfully produces a dynamic, visually rich scene adhering to the complex description.
}
\vspace{-13pt}
\label{fig:t2v_qualify}
\end{figure*}

\subsection{Implementation Details}
\paragraph{Benchmark.}
We perform comprehensive evaluations across three tasks: class-conditional image generation, text-to-image (T2I) and text-to-video (T2V) generation.
For class-conditional generation, we use ImageNet at a $256 \times 256$ resolution. 
For T2I evaluation, we use two popular benchmarks: HPSv2.1 \citep{wu2023human}, a benchmark designed to evaluate alignment with human preferences across 3,200 prompts in four styles, and T2I-CompBench \citep{huang2023t2i} for assessing performance in complex scenes.
In addition to the benchmark-specific evaluation metrics, we employ Qalign \citep{wu2023q} to compute aesthetic scores for a more comprehensive assessment. 
For T2V evaluation \citep{liu2024survey,ling2025vmbench,feng2025narrlv,chen2025finger}, we adopt the standard prompts and evaluation metrics provided by VBench \citep{huang2024vbench}.

\paragraph{Baselines.}

For T2I task, we employ the high-performing Stable Diffusion 3 (SD3) \citep{sd3} and SD3.5 \citep{sd3.5}. For T2V task, we utilize the latest Wan-1.3B and Wan-14B models \citep{wan2025wan}. Furthermore, to demonstrate the versatility of our guidance approach, we conduct a comparative analysis not only against original CFG but also with five state-of-the-art methods: CFG++~\citep{cfg++}, CFG-Zero~\citep{cfgzero}, APG~\citep{apg}, STG~\citep{stg} and SEG \citep{seg}. 
See Appendix \ref{supp:flux} for additional evaluations and Appendix \ref{supp:details} for implementation details.

\begin{wraptable}[10]{r}{0.35\textwidth} 
  
  \vspace{-32pt} 
  
  \centering
  
  \setlength{\tabcolsep}{4pt} 
  
  \renewcommand{\arraystretch}{0.9} 
  
  \begin{tabular}{@{}lcc@{}} 
    \toprule
    Method & IS$\uparrow$ & FID$\downarrow$ \\
    \midrule
    Baseline          & 125.13 & 9.41 \\
    w/ CFG            & 258.09 & 2.15 \\
    w/ ADG            & 257.92 & 2.37 \\
    w/ CFG++          & 257.04 & 2.25 \\
    w/ SEG            & 258.35 & 2.29 \\
    w/ CFG-Zero       & 258.87 & 2.10 \\
    \midrule
    \textbf{w/ Ours} & \textbf{259.12} & \textbf{2.03} \\
    \bottomrule
  \end{tabular}
  
  \captionsetup{skip=4pt} 
  \caption{\textbf{Quantitative evaluation on ImageNet $256 \times 256$ dataset}.}
  \label{tab:imagenet}
  
  \vspace{-15pt} 
\end{wraptable}

\begin{table*}[t!]
\centering

\resizebox{\linewidth}{!}{
\setlength{\tabcolsep}{6pt}

\begin{tabular}{l|l|*{9}{c}} 
\toprule
\multicolumn{1}{c|}{\multirow{2}{*}{Model}} & \multicolumn{1}{c|}{\multirow{2}{*}{Method}} 
& Total       & Quality     & Semantic    & Subject     & Background  & Aesthetic   & Imaging      & Object      & Appearance \\
& & Score       & Score       & Score       & Consistency & Consistency & Quality     & Quality    & Class       & Style      \\ 
\midrule
\multirow{6}{*}{Wan1.3B} 
& CFG       & 80.29           & 84.32   & 64.16          & 96.53         & 95.46          & \textbf{60.52} & 67.65         & 77.06          & 20.15 \\
& CFG++     & 80.35           & 83.58          & \textbf{67.43}  & \textbf{96.70}  & 93.28          & 59.02          & \textbf{69.14} & 70.06          & 19.75 \\
& APG       & 70.83           & 77.13          & 45.61          & 96.45          & 95.39          & 49.42          & 64.39          & 59.02          & 20.01 \\
& STG       & 78.78           & 83.92          & 58.19          & 95.03          & \textbf{96.04} & 59.03          & 65.59          & 68.20          & \textbf{21.51} \\
& CFG-Zero  & \underline{80.71} & \underline{84.51} & 65.53         & 96.33          & 94.56          & \underline{59.69} & \underline{69.05}          & \textbf{78.16}          & 20.31 \\
& \textbf{Ours} & \textbf{80.93} & \textbf{84.74} & \underline{65.70}          & \underline{96.57} & \underline{95.80} & \textbf{60.52} & 68.19          & \underline{78.09}          & \underline{20.59} \\
\midrule
\multirow{2}{*}{Wan14B} 
& CFG       & 82.65           & 84.88          & 73.76          & \textbf{94.45}  & \textbf{97.66}          & 68.68          & \textbf{67.82}          & 84.97          & 22.14 \\
& \textbf{Ours} & \textbf{82.84} & \textbf{84.89} & \textbf{74.65} & 94.21 & 97.56 & \textbf{68.78} & 67.77 & \textbf{89.08} & \textbf{22.27} \\
\bottomrule
\end{tabular}
}
\caption{\textbf{Quantitative comparison on VBench.}
    \ours{} consistently outperforms mainstream methods on both Wan-1.3B and Wan-14B models.
    While evaluated on all 16 dimensions, this table shows a representative subset of 9 key metrics.
    Our method achieves the highest \textbf{Total Score} and demonstrates significant improvements. 
    Best results are in \textbf{bold}; second-best are \underline{underlined}. 
}
\vspace{-15pt}
\label{tab:vbench_final_highlight_zjs}
\end{table*}

\subsection{Class-Conditional ImageNet Generation}
Evaluated on ImageNet $256 \times 256$ with a pretrained SiT-XL model \citep{sit}, \ours demonstrates clear superiority over both CFG and other advanced guidance strategies (many of which are not designed for advanced flow-based models and thus struggle to perform well \citep{cfgzero}). As shown in Table~\ref{tab:imagenet}, our method achieves the best performance, attaining both the highest Inception Score of \textbf{259.22} for image diversity and fidelity, and the lowest FID of \textbf{2.08} for perceptual quality and distributional alignment.

\subsection{Text-to-Image Generation}
The quantitative comparisons are presented in
Table \ref{tab:hpsv2_t2i_full_fixed}.
On HPSv2.1, \ours achieves the best performance not only in average scores but also across all individual dimensions, demonstrating the effectiveness of our method. By steering the sampling trajectory away from suboptimal paths inherent in CFG, \ours achieves better alignment with human preferences.
The performance on T2I-CompBench further highlights the strength of our approach, showcasing its effectiveness in handling complex generation tasks. Moreover, the high aesthetic scores confirm our method's ability to produce images with superior visual appeal.

The qualitative comparisons are presented in Figure~\ref{fig:t2i_qualify}.
Compared to CFG and other methods, \ours achieves significant improvements in both visual quality and semantic coherence: it produces higher-quality images with finer details and better semantic alignment with text descriptions, a result consistent with our toy examples.



\subsection{Text-to-Video Generation}
The quantitative comparisons are presented in Table \ref{tab:vbench_final_highlight_zjs}.
On the Wan-1.3B model, \ours achieves the highest \textbf{Total Score} (80.93), outperforming all baselines. 
We further conduct experiments on the larger Wan-14B model, demonstrating significant improvements compared to CFG.

The quantitative comparisons are presented in Figure \ref{fig:t2v_qualify}.
Our method generates videos with substantially improved quality and coherence. 
The examples highlight that \ours{} effectively addresses two critical failures of original CFG: the loss of physical plausibility in object motion and the inability to adhere to complex, compositional prompts.
Consequently, our approach yields videos that are not only more physically realistic but also demonstrate superior prompt coherence, faithfully realizing the user's creative intent.

\begin{figure}[t]
\centering
\includegraphics[width=1.0\columnwidth]{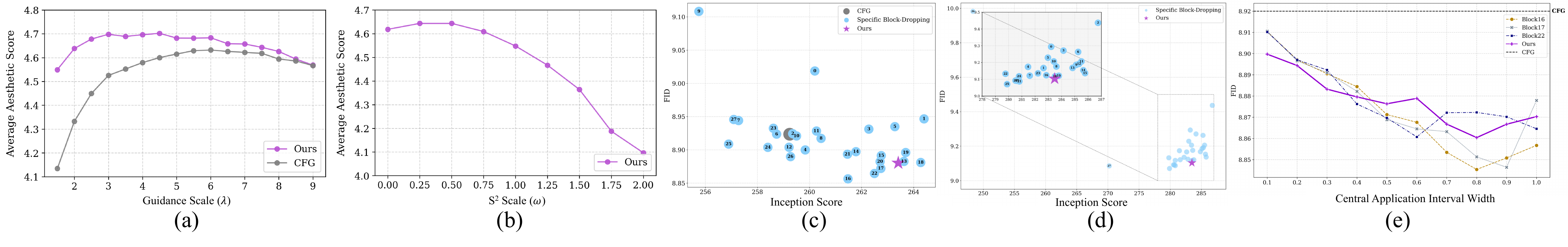}
\caption{
    \textbf{Comprehensive ablation analysis of \ours.}
    \textbf{(a)} Comparing aesthetic scores of \ours and CFG across various guidance scales ($\lambda$).
    \textbf{(b)} Analyzing aesthetic scores of \ours across various self-guidance scales ($\omega$).
    \textbf{(c, d)} Comparison of our stochastic block-dropping strategy against dropping a single, fixed block for the SiT and DiT architectures, respectively. Performance is measured by the FID-IS trade-off, where the lower-right corner indicates a better balance.
    \textbf{(e)} Ablation on the application range of \ours. The x-axis represents the width of the central interval of of noise levels where block-dropping is applied (e.g., a value of 0.2 corresponds to the central 20\% of the denoising process).
}

\label{fig:ablation_scale}
\vspace{-15pt}
\end{figure}

We further perform user study for both T2I and T2V generation. Our method is significantly preferred over all baselines in terms of both visual quality and prompt alignment. Full details are presented in Appendix \ref{supp:user}.

\subsection{Ablation Study}
\label{exp:ab_study}

\paragraph{Performance across Different Guidance Scales.}
\label{para:strategy}
We conduct experiments to compare \ours with CFG across various guidance scales, focusing on aesthetic scores. As shown in Figure~\ref{fig:ablation_scale} (a), \ours consistently outperforms CFG across a wide range of scales. Unlike CFG, which shows significant performance variance depending on the scale, our method exhibits stability and achieves high performance with minimal sensitivity to guidance scales. 
Notably, in most cases, our method even surpasses the best performance achieved by CFG, demonstrating its robustness.

\paragraph{Analysis of Block-dropping Strategy. }
We conduct a series of experiments to thoroughly analyze the effectiveness and robustness of our block-dropping strategy. 
First, to investigate the \textbf{importance of individual block}, we perform experiments on diverse model architectures.
We drop a single, specific block throughout the entire denoising process to obtain the sub-network prediction and compare its performance against our method.
As shown in Figure~\ref{fig:ablation_scale} (c,d), dropping the initial block consistently leads to performance degradation across both models. 
However, for the remaining blocks, this block-wise ablation does not yield a universal rule. We find that the optimal block to drop varies significantly across different architectures, a challenge also highlighted by AutoGuidance \citep{Autoguidance}. 
In contrast, our method eliminates the need for such complex tuning. Its simple stochastic strategy automatically outperforms most meticulously selected fixed configurations.
Furthermore, inspired by \citep{intervalcfg}, we analyze the \textbf{optimal application interval} for block-dropping. 
As shown in the Figure \ref{fig:ablation_scale} (e), applying block-dropping within the central 80\% interval of noise levels yields robust performance. 
Our method reduces FID compared to CFG and often outperforms the top-performing configurations derived from prior block-wise ablation.



\paragraph{Effect of \( S^2 \) Scale \( \omega \).}
We conduct experiments to analyze the scale of \ours \( \omega \) , as shown in Figure \ref{fig:ablation_scale} (b). When \( \omega \) is set to a smaller value, it improves the aesthetic score. However, since CFG has already produced a suboptimal result, using a larger \( \omega \) tends to overadjust, leading to a decline in quality.

\paragraph{Analysis of \ours and Naive \ours.}
\label{supp:naive_compare}
We compare applying block-dropping once versus multiple times per sampling step. Empirically, increasing the number of applications yields diminishing returns in aesthetic scores (see Figure~\ref{fig:repeat} in Appendix). We therefore conclude that a single application per timestep is sufficient, striking an effective balance between high performance and computational efficiency.
See Appendix \ref{ssec:theory} for the theoretical analysis and Appendix \ref{sssec:trace} for further visualizations.

\paragraph{Effect of Drop-Ratio.}
\label{supp:ratio}
We investigate the impact of the drop-ratio on the SD3.5 model with 24 blocks. As shown in Table~\ref{fig:ablation_drop_ratio}, when the number of dropped blocks is limited to 3/24 (approximately 10\%), the aesthetic score remains stable at a relatively high level.
However, dropping more blocks leads to a gradual decline in performance.
Empirically, we observe that a drop-ratio of about 10\% significantly improves performance.

\begin{wraptable}[12]{r}{0.35\textwidth} 
  \vspace{-10pt} 
  \centering
  \begin{tabular}{@{}cc@{}}
      \toprule
      \multicolumn{2}{c}{Stochastic Block-Dropping} \\ \midrule
      Num. & Aes. \\ \midrule
      0     & 4.618 \\
      1     & 4.652 \\
      2     & 4.643 \\
      3     & 4.616 \\
      4     & 4.531 \\ \bottomrule
  \end{tabular}
  \caption{Effect of the number of dropped blocks on aesthetic scores.}
  \label{fig:ablation_drop_ratio}
\end{wraptable}

\paragraph{Visual Analysis of Block-Dropping Impact}
To intuitively address concerns about how dropping blocks affects the final output, we provide a visual analysis in Figure~\ref{fig:vis_drop_block} (Appendix \ref{supp:flux}). This figure presents the results of an extreme test case on the SiT-XL model for the ImageNet 256$\times$256 task. In this setup, for each of the 28 generated images, a single, fixed transformer block was dropped for the entire duration of the inference process. As can be observed, the resulting images exhibit remarkable visual consistency and coherence, with no single dropped block leading to severe artifacts or a collapse in quality. This provides compelling visual evidence of the model's inherent robustness against block-level perturbations.

\paragraph{Computational Cost and Peak Memory}
We conduct a direct comparison of FLOPs, runtime, and peak memory requirements against standard CFG. The benchmark, performed on a text-to-image task with 28 inference steps, is summarized in Table~\ref{tab:cost}. The results show that our \ours incurs an overhead of approximately 40\% in both runtime and computational cost. While this entails a notable overhead, we posit that it is justified by a superior performance-efficiency trade-off, as we demonstrate in the subsequent analysis. Notably, peak GPU memory allocation remains unchanged. This is because the two forward passes within each denoising step—one for the full model and one for the sub-network—are executed sequentially. The memory from the first pass is released before the second begins, ensuring the peak memory footprint does not exceed that of a single standard CFG evaluation.

\begin{table}[!htbp]
\centering
\small 
\setlength{\tabcolsep}{4pt}
\begin{tabular}{lccc}
\toprule
\textbf{Method} & \textbf{Total Runtime (s)} & \textbf{Transformer FLOPs (TFLOPs)} & \textbf{Peak GPU Memory (GB)} \\
\midrule
CFG   & 29.2 & 168.4 & $\sim$33.8 \\
\textbf{Ours}  & 40.2 & 237.6 & $\sim$33.8 \\
\bottomrule
\end{tabular}
\caption{Computational cost and memory comparison.}
\label{tab:cost}
\end{table}

\paragraph{Performance-Efficiency Trade-off}
While our method entails a computational overhead, we argue it is justified by a superior performance-efficiency trade-off, as analyzed in Figure~\ref{fig:hps_vs_steps} (Appendix \ref{supp:flux}). This figure plots the HPS Score against a normalized computational cost, where the cost for \ours is scaled by a factor of 1.4 to account for its $\sim$40\% overhead per step. The results clearly show that our method establishes a more favorable performance-efficiency frontier, consistently achieving higher performance for any given computational budget. For instance, \ours with only 20 inference steps (equivalent cost of 28) surpasses the HPS score of standard CFG with 60 steps. This analysis compellingly demonstrates that our approach is a more practical and advanced choice for maximizing generation quality within a given computational budget.

%% file: sec/5_conclusion.tex
\section{Conclusion}

In this work, we propose \ours, a training-free stochastic self-guidance method that enhances diffusion transformers by improving the CFG mechanism.
We first conduct an empirical analysis of CFG, revealing that it often generates suboptimal results. Building on this insight, we introduce \ours, which leverages stochastic block-dropping during the forward pass to effectively guiding the model away from potential low-quality predictions, thereby improving fidelity.
Theoretical analysis and extensive experiments, including class-conditional image, text-to-image and text-to-video generation across multiple models and benchmarks, demonstrate that \ours delivers superior performance, consistently surpassing CFG and other advanced guidance strategies.

%% file: sec/6_appendix.tex
\appendix
\setcounter{page}{1}
 
\maketitle

\section*{Overview}

This appendix is divided into three main parts, covering method details, experimental supplements, and future work.

\paragraph{Appendix A}  
The first part focuses on the details and discussions of the methodology, including:
\begin{itemize}
    \item A principled derivation of Naive \ours from a Bayesian perspective.
    \item A detailed analysis between \ours and Naive \ours.
\end{itemize}

\paragraph{Appendix B}  
The second part provides supplementary information about experiments, covering: 
\begin{itemize}
    \item Explanation of the toy example and additional experimental results.
    \item More comprehensive evaluation and ablation study.
    \item User study.
    \item Further implementation details of the experiments.
    \item Detailed Prompts for the Experiments.

\end{itemize}

\paragraph{Appendix C}

The third part briefly discusses future work and potential applications.

\section{Extended Discussion and Analysis of Our Methods}
\label{app:methods}

\subsection{A Principled Derivation of Naive \ours from a Bayesian Perspective}
\label{ssec:theory}
In this subsection, we provide a principled theoretical foundation for our proposed Naive Stochastic Sub-network Guidance (Naive \ours) method. We move beyond a heuristic interpretation and formally derive our approach by drawing a direct line to the principles of Bayesian inference, as established in the seminal work ``Dropout as a Bayesian Approximation" by \citet{dropout}. Our central argument is that Naive \ours is not merely inspired by Bayesian ideas, but can be derived as a principled mechanism for correcting the predictions of a deterministic model by leveraging its own epistemic uncertainty.

\subsubsection{Foundational Bayesian Formulation}
Let $\mathcal{D} = \{x_i, c_i\}_{i=1}^M$ be our training dataset. A fully Bayesian approach to generative modeling would seek to compute the true posterior predictive distribution for a new sample $x_t$ given a condition $c$:
\begin{equation}
    p(D | x_t, c, \mathcal{D}) = \int_{\boldsymbol{\Theta}} p(D | x_t, c, \boldsymbol{\theta}) p(\boldsymbol{\theta} | \mathcal{D}) d\boldsymbol{\theta},
    \label{eq:true_posterior_pred}
\end{equation}
where $\boldsymbol{\theta} \in \boldsymbol{\Theta}$ are the model parameters, $p(\boldsymbol{\theta} | \mathcal{D})$ is the true posterior distribution over these parameters, and $p(D | x_t, c, \boldsymbol{\theta})$ is the likelihood of a specific prediction $D$ given parameters $\boldsymbol{\theta}$. The true posterior is given by Bayes' theorem:
\begin{equation}
    p(\boldsymbol{\theta} | \mathcal{D}) = \frac{p(\mathcal{D} | \boldsymbol{\theta}) p(\boldsymbol{\theta})}{p(\mathcal{D})} = \frac{p(\mathcal{D} | \boldsymbol{\theta}) p(\boldsymbol{\theta})}{\int_{\boldsymbol{\Theta}} p(\mathcal{D} | \boldsymbol{\theta}') p(\boldsymbol{\theta}') d\boldsymbol{\theta}'}.
\end{equation}
The integral in the denominator, known as the marginal likelihood or model evidence, is intractable for deep neural networks. To circumvent this, we employ Variational Inference (VI), introducing a tractable approximate posterior distribution $q_{\phi}(\boldsymbol{\theta})$ (parameterized by $\phi$) to approximate $p(\boldsymbol{\theta} | \mathcal{D})$. We minimize the Kullback-Leibler (KL) divergence between these two distributions:
\begin{align}
    \phi^* &= \arg\min_{\phi} \text{KL}(q_{\phi}(\boldsymbol{\theta}) || p(\boldsymbol{\theta} | \mathcal{D})) \\
    &= \arg\min_{\phi} \int q_{\phi}(\boldsymbol{\theta}) \log \frac{q_{\phi}(\boldsymbol{\theta})}{p(\boldsymbol{\theta} | \mathcal{D})} d\boldsymbol{\theta} \\
    &= \arg\min_{\phi} \int q_{\phi}(\boldsymbol{\theta}) \log \frac{q_{\phi}(\boldsymbol{\theta}) p(\mathcal{D})}{p(\mathcal{D} | \boldsymbol{\theta}) p(\boldsymbol{\theta})} d\boldsymbol{\theta} \\
    &= \arg\min_{\phi} \left( \text{KL}(q_{\phi}(\boldsymbol{\theta}) || p(\boldsymbol{\theta})) - \mathbb{E}_{q_{\phi}(\boldsymbol{\theta})}[\log p(\mathcal{D}|\boldsymbol{\theta})] \right). \label{eq:elbo_min}
\end{align}
Minimizing this objective is equivalent to maximizing the Evidence Lower Bound (ELBO), $\mathcal{L}_{\text{ELBO}}$. 
The work of \citet{dropout} provides the theoretical grounding for interpreting stochastic network perturbations, such as dropout, as a form of this Bayesian optimization.

In our work, we generalize this concept from neuron-level dropout to block-level dropout. Each binary mask $\mathbf{m}_i \sim p(\mathbf{m})$ applied via stochastic block dropping effectively samples a specific set of weights $\boldsymbol{\theta}_i = \boldsymbol{\theta} \odot \mathbf{m}_i$ from this approximate posterior, which we denote simply as $q(\boldsymbol{\theta})$.

\subsubsection{Monte Carlo Estimation of the Approximate Posterior Predictive}

The prediction of a single sub-network, $\hat{D}_\theta(x_t \mid c, \mathbf{m}_i)$, is a legitimate sample from the \textit{approximate posterior predictive distribution}, $p_q(D | x_t, c)$:
\begin{equation}
    \hat{D}_\theta(x_t \mid c, \mathbf{m}_i) \triangleq D(x_t \mid c; \boldsymbol{\theta}_i), \quad \text{where} \quad \boldsymbol{\theta}_i \sim q(\boldsymbol{\theta}).
\end{equation}

The first moment of this distribution, the posterior mean $\mu_{\text{post}}$, is theoretically defined as the integral over the variational distribution:
\begin{equation}
    \mu_{\text{post}}(x_t \mid c) \triangleq \mathbb{E}_{q(\boldsymbol{\theta})}[D(x_t \mid c; \boldsymbol{\theta})] = \int D(x_t \mid c; \boldsymbol{\theta}) q(\boldsymbol{\theta}) d\boldsymbol{\theta}.
    \label{eq:posterior_mean_def}
\end{equation}

Since this integral is analytically intractable for deep neural networks, we rely on \textbf{Monte Carlo integration} to estimate it. The empirical average computed by our algorithm serves as this estimator:
\begin{equation}
    \hat{\mu}_{\text{post}}(x_t \mid c) \approx \frac{1}{N} \sum_{i=1}^N \hat{D}_\theta(x_t \mid c, \mathbf{m}_i).
    \label{eq:posterior_mean_est}
\end{equation}

Computing high-dimensional integrals via empirical averaging over diverse predictive hypotheses is a standard practice in deep learning. This paradigm is supported by extensive literature, including explicit methods like \cite{deepensembles, snapshot, dropout,lei2023masked, batchensemble}. These works collectively establish that aggregating predictions from stochastic sub-states or ensemble members effectively approximates the predictive posterior. Our algorithm is a direct application of this principle to the sub-networks induced by block-dropping.

This posterior mean, $\mu_{\text{post}}$, represents the ``center of mass" of the model's belief. The second central moment, the variance, quantifies the \textbf{epistemic uncertainty}:
\begin{align}
    \text{Var}_{q(\boldsymbol{\theta})}[D(x_t \mid c; \boldsymbol{\theta})] 
    &= \mathbb{E}_{q(\boldsymbol{\theta})}\big[(D(x_t ; \boldsymbol{\theta}) - \mu_{\text{post}})^2\big] \notag \\
    &\approx \frac{1}{N} \sum_{i=1}^N \big( \hat{D}_\theta(x_t; \mathbf{m}_i) - \hat{\mu}_{\text{post}}(x_t) \big)^2.
\end{align}

Our central hypothesis is that \textbf{low-quality generative outputs often arise in regions of high epistemic uncertainty}. In such regions, the posterior mean, $\mu_{\text{post}}$, often corresponds to a ``safe," but ultimately low-quality output (e.g., a blurry artifact). The deterministic MAP estimate, $D_\theta(x_t \mid c)$, however, might be unjustifiably confident in these very regions.

\subsubsection{Deriving \ours as an Uncertainty-Aware Correction}
Based on this hypothesis, we formulate a principled correction to the Classifier-free Guidance (CFG) prediction, $\tilde{D}_{\text{CFG}}$. The standard guidance is:
\begin{equation}
    \tilde{D}_{\text{CFG}}(x_t \mid c) = D_\theta(x_t \mid \phi) + \lambda \big( D_\theta(x_t \mid c) - D_\theta(x_t \mid \phi) \big).
\end{equation}
We define our corrected prediction, $\tilde{D}_\theta^{\lambda, \omega}(x_t \mid c)$, as the solution to an optimization problem where we seek a prediction that remains faithful to the original guidance while being repelled from the center of uncertainty. Let us define a correction vector $\Delta D$. We propose that this correction should be in the direction opposite to the posterior mean, which acts as the locus of uncertainty-induced artifacts:
\begin{equation}
    \Delta D \triangleq - \omega \cdot \mu_{\text{post}}(x_t \mid c),
\end{equation}
where $\omega$ is a scalar controlling the magnitude of the repulsion. The corrected prediction is thus the linear superposition of the original guidance and this correction term:
\begin{align}
    \tilde{D}_\theta^{\lambda, \omega}(x_t \mid c) & \triangleq \tilde{D}_{\text{CFG}}(x_t \mid c) + \Delta D, \label{eq:principled_derivation_step1} \\
    & = \tilde{D}_{\text{CFG}}(x_t \mid c) 
        - \omega \cdot \mu_{\text{post}}(x_t \mid c), \label{eq:corrected_guidance} \\
    & = \underbrace{D_\theta(x_t \mid \phi) 
        + \lambda \big( D_\theta(x_t \mid c) - D_\theta(x_t \mid \phi) \big)}_{\text{Standard CFG}} \notag \\
    & \quad - \underbrace{\omega \cdot \mathbb{E}_{q(\boldsymbol{\theta})}[D(x_t \mid c; \boldsymbol{\theta})]}_{\text{Uncertainty-Aware Repulsion Term}}. \label{eq:principled_derivation_step2}
\end{align}


Substituting the Monte Carlo approximation from Eq. \ref{eq:posterior_mean_est} into Eq. \ref{eq:principled_derivation_step1}, we recover our full Naive \ours formulation:
\begin{align}
    \tilde{D}_\theta^{\lambda, \omega}(x_t \mid c) &= D_\theta(x_t \mid \phi) 
    + \lambda \big( D_\theta(x_t \mid c) - D_\theta(x_t \mid \phi) \big) \notag \\
    &\quad - \frac{\omega}{N} \sum_{i=1}^N \hat{D}_\theta(x_t \mid c, \mathbf{m}_i).
    \label{eq:final_s2g_recovered}
\end{align}

\subsubsection{Theoretical Interpretation and Decompositions}
This derivation provides a much deeper understanding of Naive \ours.

\paragraph{Decomposition of Predictive Components.} We can rearrange Eq. \ref{eq:principled_derivation_step2} to analyze the contribution of each component to the final prediction:
\begin{align}
    \tilde{D}_\theta^{\lambda, \omega}(x_t \mid c) 
    &= (1-\lambda) D_\theta(x_t \mid \phi) 
    + \lambda D_\theta(x_t \mid c) \notag \\
    &\quad - \omega \cdot \mu_{\text{post}}(x_t \mid c) \\
    &= \underbrace{\lambda D_\theta(x_t \mid c)}_{\text{MAP Guidance}} 
    + \underbrace{(1-\lambda) D_\theta(x_t \mid \phi)}_{\text{Unconditional Prior}} \notag \\
    &\quad - \underbrace{\omega \cdot \mu_{\text{post}}(x_t \mid c)}_{\text{Bayesian Correction}}.
\end{align}

This shows a clear trade-off: we leverage the strong guidance from the conditional MAP estimate ($D_\theta(x_t \mid c)$) and the unconditional prior ($D_\theta(x_t \mid \phi)$), but temper both with a Bayesian correction term that represents the consensus of a diverse committee of model hypotheses. It acts to regularize the overconfidence of the single MAP estimate.

\paragraph{A Gradient-Space Perspective.} In diffusion models, the guidance is applied in the noise prediction space. Let $\epsilon_\theta(x_t, c)$ be the model's noise prediction. The standard CFG-guided noise $\tilde{\epsilon}_{\text{CFG}}$ is:
\begin{equation}
    \tilde{\epsilon}_{\text{CFG}}(x_t, c) = \epsilon_\theta(x_t, \phi) + \lambda (\epsilon_\theta(x_t, c) - \epsilon_\theta(x_t, \phi)).
\end{equation}
Our method introduces a correction term directly in this space. Let $\bar{\epsilon}_{\text{post}}(x_t, c) = \mathbb{E}_{q(\boldsymbol{\theta})}[\epsilon_\theta(x_t, c; \boldsymbol{\theta})]$ be the posterior mean of the noise prediction. Our corrected noise prediction becomes:
\begin{align}
    \tilde{\epsilon}_{\text{S}^2\text{G}}(x_t, c) &= \tilde{\epsilon}_{\text{CFG}}(x_t, c) - \omega \cdot \bar{\epsilon}_{\text{post}}(x_t, c) \\
    &= \tilde{\epsilon}_{\text{CFG}}(x_t, c) - \omega \cdot \left(\frac{1}{N}\sum_{i=1}^N \epsilon_\theta(x_t, c; \boldsymbol{\theta}_i)\right).
\end{align}
This reveals that Naive \ours is performing a direct modification of the guidance vector at each step of the denoising process. The repulsion from the ``center of uncertainty" is not an abstract concept but a concrete vector subtraction in the high-dimensional noise space.

\paragraph{Connection to Negative Ensemble Distillation.} Our method can be framed as a novel form of \textit{negative distillation} applied at inference time. Standard ensemble distillation trains a single model to mimic the average output of an ensemble. In contrast, Naive \ours uses the ensemble's average prediction ($\mu_{\text{post}}$) not as a target to be imitated, but as an anti-target to be actively repelled. This ``distillation-rejection" mechanism is a new and principled way to harness the wisdom of an ensemble without collapsing to its mean.

In summary, Naive \ours is a theoretically grounded method that leverages the principles of Bayesian model averaging and uncertainty quantification. It operationalizes the insight that high-quality generation requires not only strong conditional guidance but also a mechanism to actively avoid regions of high model uncertainty. Our derivation shows that subtracting the Monte Carlo average of stochastic sub-networks is a direct and principled way to implement this avoidance, thereby correcting for the inherent limitations of a single, deterministic generative model, as shown in Figure~\ref{fig:toy_trajectories}.

\begin{figure}[t]
\centering
\includegraphics[width=1.0\columnwidth]{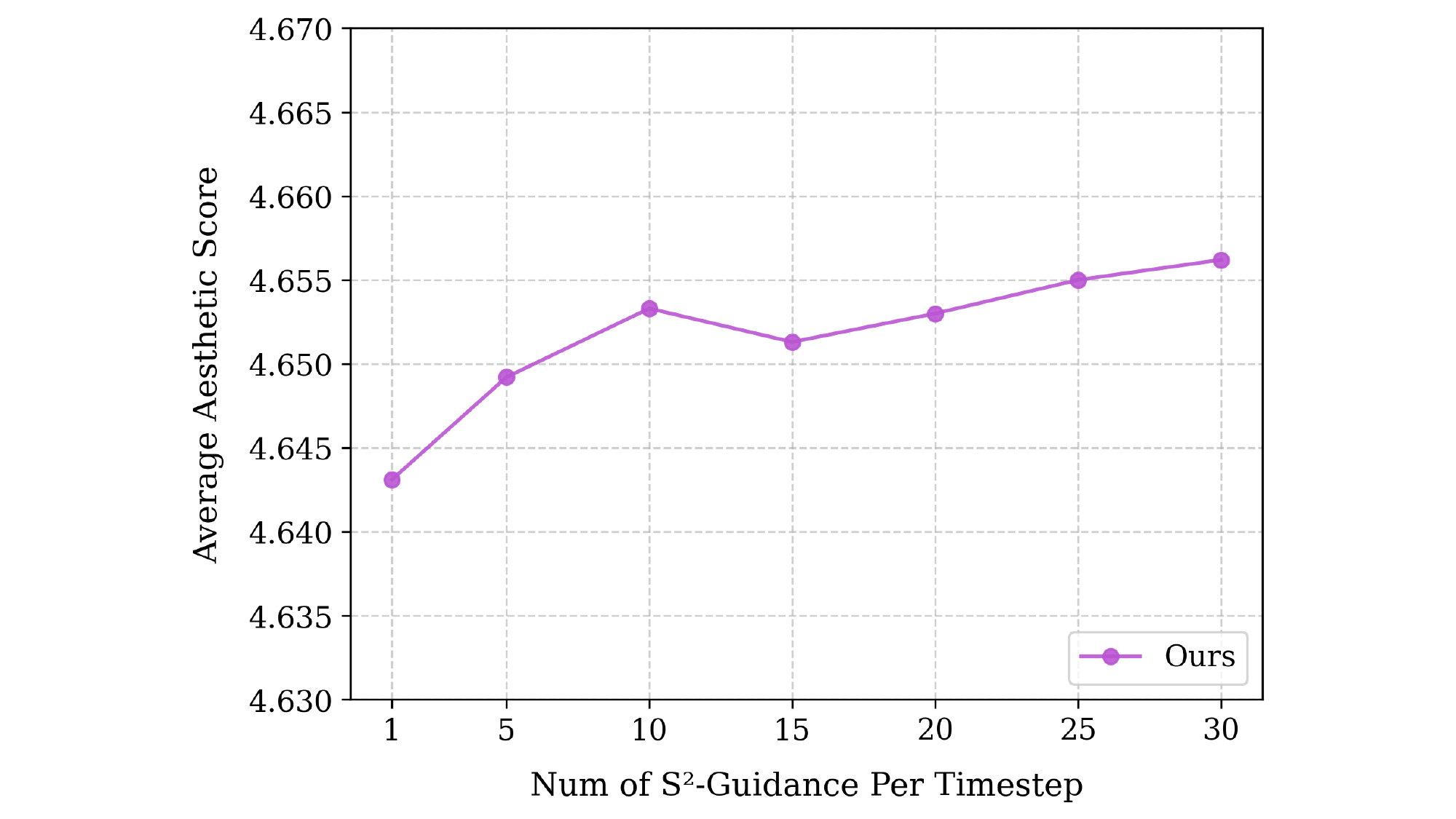}
\caption{
        \textbf{Aesthetic score gains brought by increasing the number of forward passes with stochastic block dropping at each time step.}
    }
    \label{fig:repeat}
\end{figure}

\subsection{Comparative Analysis of \ours and Naive \ours}
\label{sec:comparative_analysis}

Our investigation into the behavior of sub-networks reveals a crucial property. We find that when the stochastic block-dropping ratio is constrained within a specific range, the guidance provided by different sub-networks appears remarkably consistent. As illustrated in Figure~\ref{fig:supp_dist}, even when different blocks are dropped to form distinct sub-network configurations, their individual guidance effects on the model's output distribution exhibit a strong similarity. 

This consistent behavior motivates us to formalize the relationship between the two methods using the principle of \textbf{unbiased estimation}.

Let $\theta$ be the model parameters and $p(\mathbf{m})$ be the distribution of binary masks. Following \cite{dropout}, the stochastic block-dropping process induces a variational distribution $q(\tilde{\theta})$ over the parameter space. We define the \textbf{Theoretical Expected Guidance} (the population mean) as the exact predictive mean under this induced distribution:
\begin{equation}
    \mathcal{G}^* \triangleq \omega \cdot \mathbb{E}_{q(\tilde{\theta})}[D(x_t \mid c; \tilde{\theta})] \equiv \omega \cdot \mathbb{E}_{\mathbf{m} \sim p(\mathbf{m})}[\hat{D}_\theta(x_t \mid c, \mathbf{m})].
\end{equation}

\textbf{Naive \ours} approximates this target using a Monte Carlo average of $N$ i.i.d. samples:
\begin{equation}
    G_{\text{Naive}} = \frac{\omega}{N} \sum_{i=1}^N \hat{D}_\theta(x_t \mid c, \mathbf{m}_i).
\end{equation}
By the linearity of expectation, it holds that $\mathbb{E}[G_{\text{Naive}}] = \mathcal{G}^*$.

In contrast, our simplified \textbf{\ours} employs a stochastic guidance term from a single sample ($N=1$):
\begin{equation}
    G_{\text{\ours}} = \omega \cdot \hat{D}_\theta(x_t \mid c, \mathbf{m}_t), \quad \text{where} \quad \mathbf{m}_t \sim p(\mathbf{m}).
\end{equation}
We formally derive that $G_{\text{\ours}}$ is also an unbiased estimator of the same theoretical target $\mathcal{G}^*$:
\begin{equation}
    \mathbb{E}_{p(\mathbf{m}_t)}[G_{\text{\ours}}] = \mathbb{E}_{p(\mathbf{m}_t)}[\omega \cdot \hat{D}_\theta(x_t \mid c, \mathbf{m}_t)] = \mathcal{G}^*.
\end{equation}

Since $\mathbb{E}[G_{\text{\ours}}] = \mathbb{E}[G_{\text{Naive}}] = \mathcal{G}^*$, both methods are mathematically \textbf{unbiased Monte Carlo estimators} of the same target, differing only in variance. While $G_{\text{\ours}}$ naturally exhibits higher variance per step compared to the ensemble average $G_{\text{Naive}}$, the iterative nature of diffusion sampling effectively performs temporal integration. This smooths out the stochastic noise over the trajectory (as confirmed by our variance analysis in Appendix B), confirming that a single stochastic sample is sufficient and theoretically justified.



Further experiments, such as repeating the process with a small number of samples (as shown in Figure~\ref{fig:repeat}), corroborate this perspective by demonstrating diminishing returns, validating the practical efficiency of our approach.


\begin{figure*}[t!]
\centering
\includegraphics[width=1.0\textwidth]{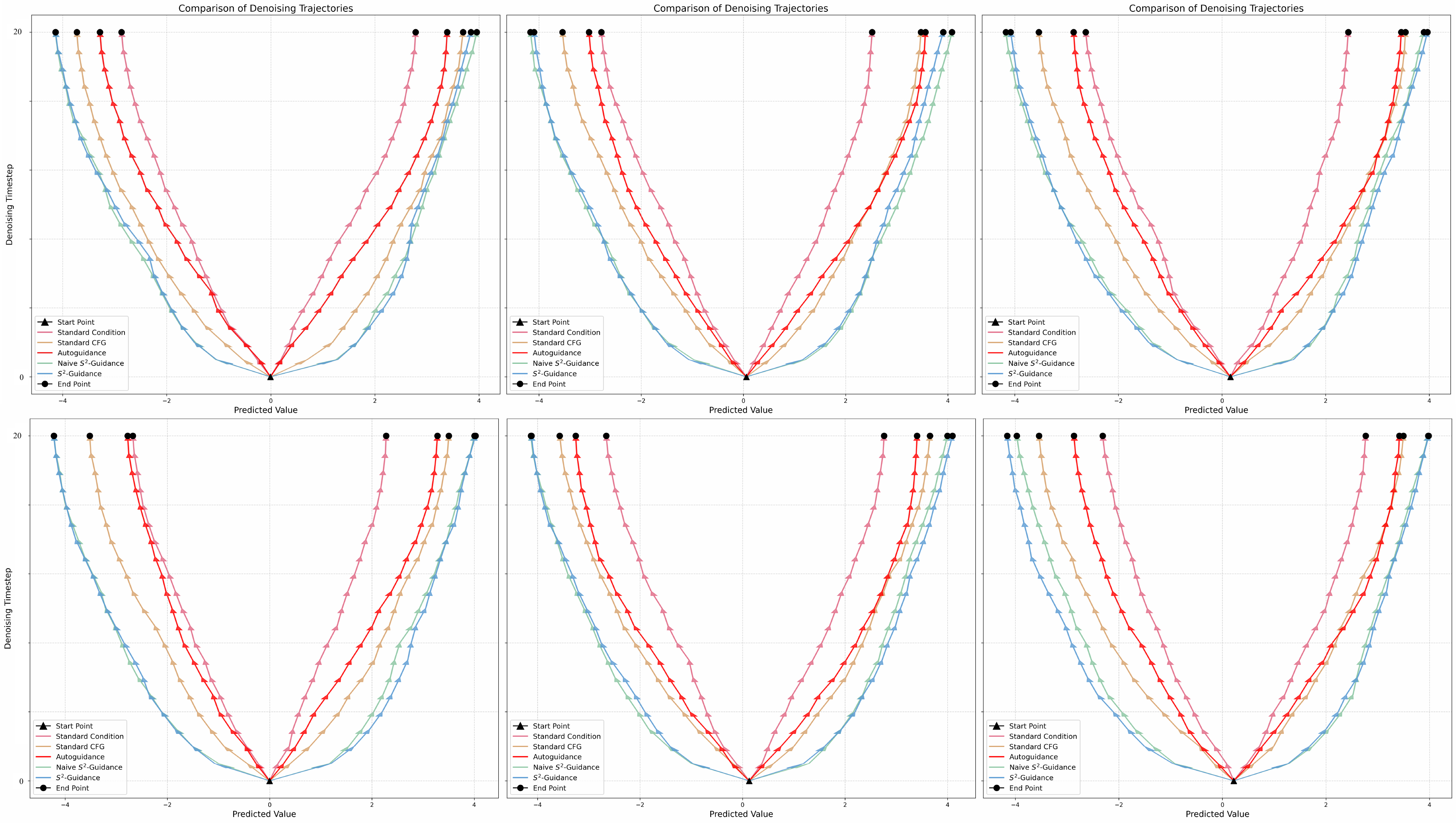}
\caption{
        \textbf{Visualization of Denoising Trajectories on the 1D Bimodal Gaussian Data.}
        Each panel shows the paths taken by different guidance methods to generate samples targeting the ground truth modes at -4 and 4. The y-axis represents the denoising timestep (from start to end), and the x-axis shows the predicted sample value. While standard CFG and Autoguidance improve upon the unguided baseline, they consistently fail to reach the ground truth. In contrast, both \textbf{Naive S²-Guidance} and our final \textbf{S²-Guidance} method successfully steer the generation process to the correct endpoints. 
        The more direct paths of our methods indicate a more accurate guidance signal throughout the entire denoising process.
    }
\label{fig:toy_trajectories}
\vspace{-5pt}
\end{figure*}

\begin{figure}[t]
\centering
\includegraphics[width=1.0\columnwidth]{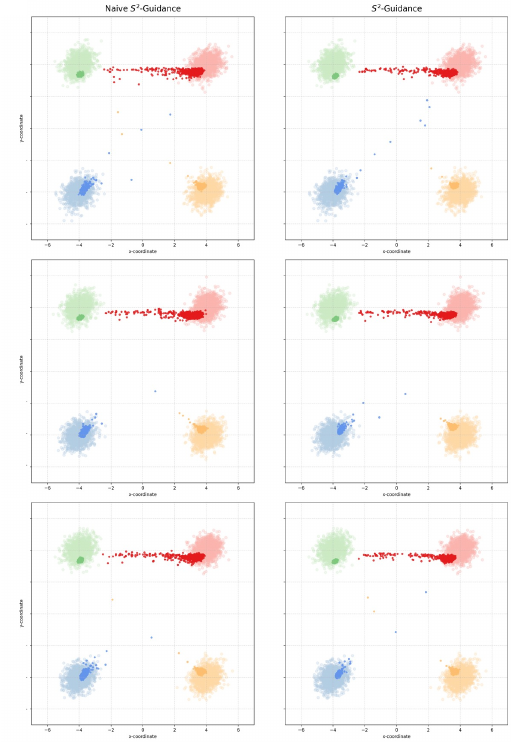}
\caption{
        \textbf{More Visual Comparisons of Naive S²-Guidance and S²-Guidance on the 2D Gaussian Mixture.}
        Left: Naive S²-Guidance. Right: S²-Guidance. Each row corresponds to a different random seed. The generated sample distributions are virtually identical, demonstrating that the performance gain from the computationally intensive naive approach is minimal. This justifies our adoption of the more efficient S²-Guidance method.
    }
    \label{fig:supp_dist}
\vspace{-5pt}
\end{figure}

\section{More Details About Our Experiments}
\label{app:experiments}

\subsection{Toy Examples}

\subsubsection{More results of Toy Examples}
\label{app:toy}

To further analyze the guidance mechanisms, we visualize the full denoising trajectories for the 1D Bimodal Gaussian Distribution in Figure ~\ref{fig:toy_trajectories}. The figure illustrates that while standard CFG and Autoguidance improve upon the unguided baseline, their final predictions consistently deviate from the distributions centered at -4 and 4. This visually demonstrates the mode-shifting problem discussed in the main paper.

In stark contrast, the paths for both Naive S²-Guidance and our final S²-Guidance method are more direct and successfully converge to the correct endpoints. This suggests that our self-guidance signal effectively corrects the generation path at each timestep, preventing the model from settling in the suboptimal regions favored by other methods.

\subsubsection{Naive \ours versus \ours in Toy Examples}
\label{sssec:trace}
In our methodology, we first proposed Naive S²-Guidance, which averages the predictions from multiple stochastic sub-networks to create a robust negative guidance signal. However, this approach carries a significant computational cost. To address this, we introduced our final, more efficient S²-Guidance, which uses only a single stochastic sub-network per timestep.

To validate that this simplification does not cause a meaningful performance degradation, we conduct a direct comparison on the 2D Gaussian mixture. As illustrated in Figure~\ref{fig:supp_dist}, the sample distributions generated by both methods are qualitatively indistinguishable across multiple independent runs. Both approaches effectively guide the generation process to the correct modes and prevent the mode collapse issues seen in standard CFG (see Figure 3 in the main paper).

Given the negligible difference in performance, the substantial computational advantage of S²-Guidance makes it the far more practical and efficient choice than the naive variant. This finding strongly supports our adoption of the simplified approach as our final method.

\subsubsection{Details of Toy Examples}
\label{sssec:toy_details}

Below are the implementation details for the experiments on both synthetic and real-world data, as referenced in the main paper. All experiments were conducted using class-balanced datasets to assess the performance of our method.

\begin{itemize}
    \item \textbf{1-D Bimodal Gaussian Distribution:}
    This experiment was designed to test the model's ability to stably and completely capture both modes of a bimodal distribution. The ground-truth data is an equally-weighted mixture of two Gaussians. The diffusion model, parameterized by a standard neural network, was trained for iterations to reconstruct the target distribution. Analysis involved visualizing the final sample distribution and denoising trajectories to show that \ours consistently covers both modes, whereas the baseline may exhibit instability or mode preference (see Figure 3 in the main paper).

    \item \textbf{2-D Gaussian Mixture (4-Modes):}
    This experiment assessed the model's capacity to generate samples from a disconnected, multi-modal manifold. The data consisted of an equally-weighted mixture of 4 isotropic Gaussians, with means located at $(-4, -4)$, $(-4, 4)$, $(4, -4)$, and $(4, 4)$. The analysis focused on the final distribution and denoising paths to demonstrate that \ours successfully captures all 4 distinct modes, improving upon the baseline's mode coverage.

    \item \textbf{Real-Image Data (CIFAR-10):}
    To validate \ours on high-dimensional data, we used a class-balanced dataset from CIFAR-10, consisting of 5,000 \textbf{'horse'} images and 5,000 \textbf{'car'} images. The diffusion model employed a neural network parameterization common for image tasks. The primary goal of the analysis was to assess the quality and class-separability of the generations. To this end, we generated 3,000 images and visualized their \textbf{CLIP (ViT-B/32) features} \c in 2-D using \textbf{t-SNE}. The resulting plot demonstrates that \ours produces more distinct and well-separated class clusters compared to the baseline, indicating higher-quality and less ambiguous generations (see Figure 4 in the main paper).
\end{itemize}

\subsection{Extended Evaluations}
\label{supp:flux}
\paragraph{Experiments using Flux.}In addition to the main experiments conducted with SD3 and SD3.5, we further evaluate our method using Flux \citep{flux}, a state-of-the-art (SOTA) model for text-to-image generation. 
Note that Flux is a CFG-distilled model, meaning that directly applying classifier-free guidance (CFG) may lead to different results. 
We use a De-distilled version of Flux~\citep{flux} in our experiments. Additionally, we follow the same benchmark setting as HPSv2.1 to ensure consistency and comparability.

\begin{table}[h!]
    \centering
    \begin{tabular}{l|ccccc|c} 
    \toprule
    \multirow{2}{*}{Method} & \multicolumn{5}{c|}{HPSv2.1(\%) $\uparrow$} & \multirow{2}{*}{Qalign$\uparrow$} \\ 
    \cmidrule(lr){2-6} 
                   & Anime & Concept Art & Paintings & Photo & Avg. & \\ 
    \midrule
    CFG  & 31.29           & 29.85                 & 30.03              & 28.16           & 29.84          & 4.65 \\
    CFG (1.4$\times$ NFE) & \textbf{31.59} & 30.10 & 30.35 & 28.47 & 30.13 & 4.68 \\
    \textbf{Ours} & 31.48  & \textbf{30.21}        & \textbf{30.48}     & \textbf{28.88}  & \textbf{30.26} & \textbf{4.70} \\
    \bottomrule
    \end{tabular}
    \caption{\textbf{Quantitative evaluation of CFG and our approach using Flux under the HPSv2.1 benchmark.}
        The HPSv2.1 grouping evaluates different styles, while Qalign measures aesthetic quality.
        Higher scores (↑) are better. Best results are in bold.
    }
    \label{tab:flux_comparison}
\end{table}

The results in Table \ref{tab:flux_comparison} show that our method consistently outperforms the baseline CFG across different categories, including Anime, Concept Art, Paintings, and Photo. Specifically, we observe an average improvement of \textbf{0.42}, highlighting the robustness and effectiveness of our approach.

For more qualitative results, please refer to Figure \ref{fig:supp_qualify_1} and Figure \ref{fig:supp_qualify_2}. These comprehensive results demonstrate the effectiveness of our proposed approach across various scenarios.

\paragraph{Analysis of Variance from Stochastic Dropping}
To assess the stability of our method, we quantify the output variance introduced by the stochastic dropping of network blocks. In our experiment, we generate multiple images for the same prompt while keeping the initial noise seed fixed, thereby isolating the variance attributable solely to the stochastic dropping process. The quantitative results, presented in Table~\ref{tab:variance}, demonstrate that the run-to-run variance is negligible. As shown, \ours exhibits a variance on the order of $10^{-6}$ and a coefficient of variation of less than 1\%. This indicates an extremely high degree of stability and output consistency, confirming that the stochastic element does not compromise the reliability of the generation process.

\begin{table}[h!]
\centering
\begin{tabular}{lcccc}
\toprule
\textbf{Method} & \textbf{Mean (\%)} & \textbf{Var.} & \textbf{Std. Dev.} & \textbf{Coeff. of Var.} \\
\midrule
CFG             & 30.48 & -- & -- & -- \\
\ours           & 30.86 & $7 \times 10^{-6}$ & 0.0026 & 0.84\% \\
\bottomrule
\end{tabular}
\caption{Analysis of variance from stochastic dropping with a fixed initial seed.}
\label{tab:variance}
\end{table}



\begin{figure}[t]
\centering
\includegraphics[width=1.0\columnwidth]{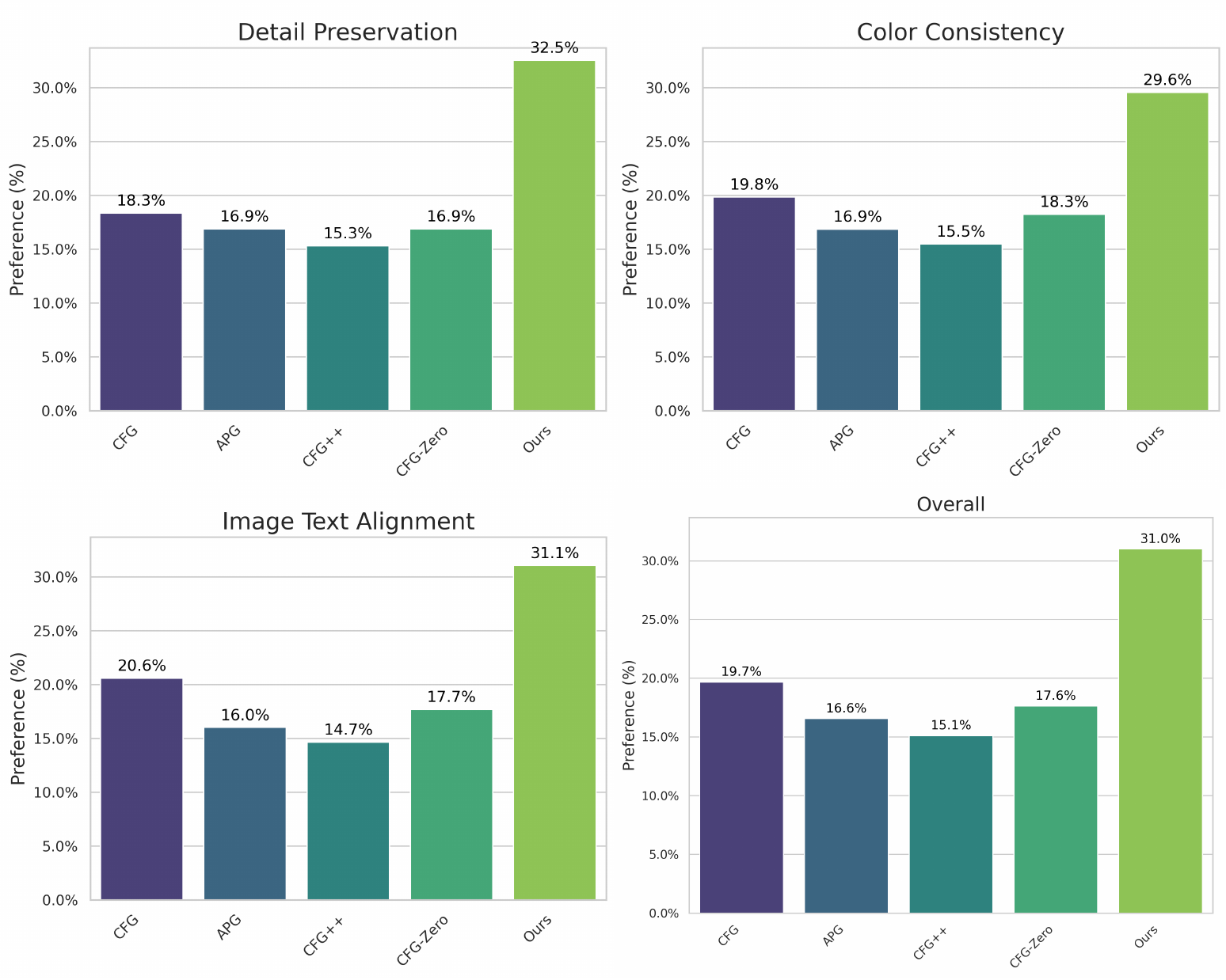}
\caption{
        \textbf{Human preference evaluation results for \ours against baseline methods.}
        The bar charts show the percentage of times each method was selected as the best for three criteria: Detail Preservation, Color Consistency, and Image-Text Alignment, along with an Overall aggregated score. Our method, \ours, is significantly preferred by human evaluators across all categories, achieving a preference rate of over 29\% in every dimension and surpassing 30\% overall. This demonstrates its robust ability to generate perceptually superior images.
    }
    \label{fig:user_study}
\vspace{-5pt}
\end{figure}

\subsection{User Study}
\label{supp:user}
To quantitatively evaluate the perceptual quality and prompt fidelity of our method, we conducted a comprehensive user study comparing \ours against four strong baselines: CFG \citep{cfg}, APG \citep{apg}, CFG++ \citep{cfg++}, and CFG-Zero \citep{cfgzero}. The evaluation was performed on images generated from a diverse set of diffusion models to assess the generalizability of our approach.

We recruited 14 participants with expertise in computer vision and generative AI. For each evaluation instance, participants were presented with a text prompt and the corresponding images generated by all five methods, displayed in a randomized order to prevent bias. Participants were instructed to evaluate the results based on three key criteria:
\begin{itemize}
    \item \textbf{Detail Preservation:} The clarity, sharpness, and richness of details in the generated image.
    \item \textbf{Color Consistency:} The naturalness, harmony, and realism of the colors.
    \item \textbf{Image-Text Alignment:} How well the generated image accurately reflects the content and intent of the text prompt.
\end{itemize}
For each criterion, participants were asked to select the image (or images) they found to be the most successful. This design allows for multiple selections if a participant deems more than one result to be of high quality for a given aspect, thereby capturing a more nuanced assessment of performance.

The results of the user study are presented in Figure~\ref{fig:user_study}. The findings demonstrate a clear and consistent preference for our proposed method, \ours, across all evaluated metrics. Specifically, in the \textit{Detail Preservation} category, \ours was preferred in 32.5\% of cases, significantly outperforming the runner-up, CFG (18.3\%). A similar dominant trend is observed for \textit{Color Consistency}, where \ours achieved a 29.6\% preference rate. Furthermore, for \textit{Image-Text Alignment}, our method was chosen 31.1\% of the time, again marking a substantial lead over all baselines.

Aggregating the votes, the \textit{Overall} preference for \ours stands at 31.0\%, confirming its comprehensive superiority. This strong performance in human evaluations validates that \ours not only improves guidance from a theoretical standpoint but also translates to tangible and perceptually superior generation quality that is easily recognized by human users.

\subsection{Implementation Details}
\label{supp:details}
To ensure fair comparisons, the implementation details of our experiments are as follows:
For the text-to-image comparisons, we used SD3 and SD3.5 \citep{sd3,sd3.5} with the guidance scale set to $7.5$. For our scale parameter $\omega$, we set it to $0.25$.
For the text-to-video comparisons, we use a guidance scale of 5.0. Similarly, our scale parameter $\omega$ is set to $0.25$.
All other hyperparameters are set to the default configurations of the respective models.
For the baseline comparisons, we follow the original implementations provided in their official repositories. Specifically, APG \citep{apg} and CFG++ \citep{cfg++} are implemented using the community-contributed versions that are integrated into the Diffusers framework.
All experiments are conducted on NVIDIA H20 GPUs with 96GB memory.

\subsection{Detailed Prompts for Figure 1 } 
\label{app:figure1_prompts}

This section provides the prompts used to generate the visual results presented in Figure 1. The examples are referenced by their grid position in the figure (row, column).

\begin{itemize}
    \item \textbf{(Top, 1) Astronaut in space (Video):} 
    \textit{“An astronaut flying in space.”}

    \item \textbf{(Top, 2) Floating Castle (Image):} 
    \textit{“A magnificent castle sitting high on a floating island above the clouds. Fluffy clouds surround the base of the island and form the text 'S2 Guidance Is All You Need' in a romantic, swirling style. The castle is adorned with towers, golden lights twinkling in the windows, and vines of blooming flowers climbing its walls. The scene is lit by a warm, golden light glowing from the sun, with a starry heaven faintly visible on the horizon.”}

    \item \textbf{(Top, 3) Abstract Portrait (Image):}
    \textit{“The bold dramatic strokes of the painter's brush created a stunning abstract masterpiece a work of emotional depth and intensity.”}

    \item \textbf{(Top, 4) Cat with Rocket (Image):}
    \textit{“A cat sitting besides a rocket on a planet with a lot of cactuses.”}

    \item \textbf{(Top, 5) Sports Car Driving (Video):}
    \textit{“a car accelerating to gain speed.”}

    \item \textbf{(Bottom, 1) Woman with Colored Powder (Video):}
    \textit{“A close-up of a beautiful woman's face with colored powder exploding around her, creating an abstract splash of vibrant hues.”}

    \item \textbf{(Bottom, 2) Woman with Umbrella (Image):}
    \textit{“A woman sitting under an umbrella in the middle of a restaurant.”}

    \item \textbf{(Bottom, 3) Man Running on Beach (Image):}
    \textit{“A man is running his hand over a smooth rock at the beach.”}

    \item \textbf{(Bottom, 4) Clay Sheep (Image):}
    \textit{“a red book and an ivory sheep.”}

    \item \textbf{(Bottom, 5) Bear Climbing Tree (Video):}
    \textit{“a bear climbing a tree.”}

\end{itemize}

\begin{figure*}[t!]
\centering
\includegraphics[width=0.98\textwidth]{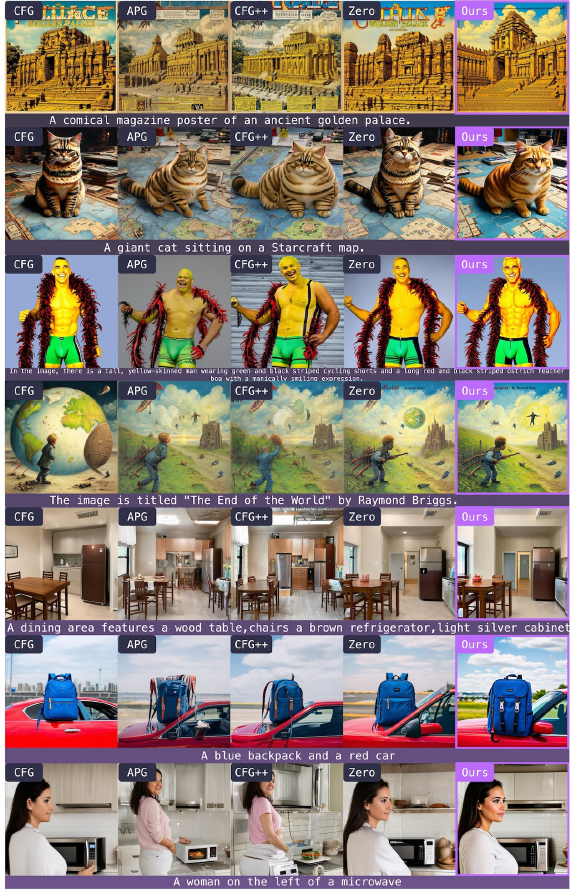}
\caption{
        \textbf{Qualitative comparison of \ours with baseline methods.}
        Our method consistently generates images with superior visual quality, better prompt alignment, and fewer artifacts across a variety of prompts.
        For instance, \ours excels at stylistic replication (row 4), complex concept combinations (row 5).
    }
\label{fig:supp_qualify_1}
\vspace{-10pt}
\end{figure*}

\begin{figure*}[t!]
\centering
\includegraphics[width=0.98\textwidth]{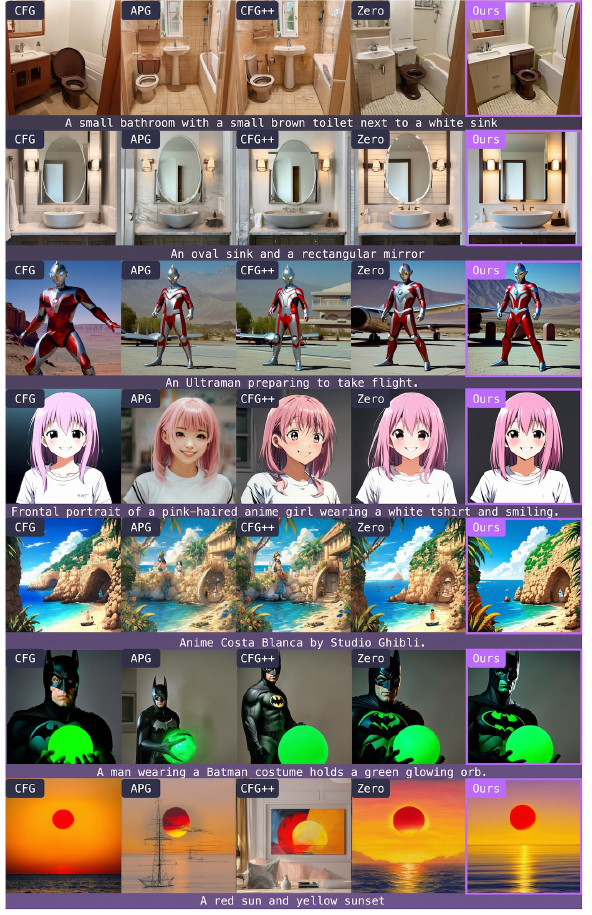}
\caption{
        \textbf{Further qualitative comparisons.}
        Our approach demonstrates robust improvements in both prompt fidelity and aesthetic quality.
        Key advantages include accurate attribute binding (e.g., ``oval sink and rectangular mirror" in row 2), faithful character and style generation (rows 3-5), and superior handling of lighting and composition (rows 6, 7).
        \ours consistently avoids the conceptual blending and visual artifacts that affect other methods.
    }
\label{fig:supp_qualify_2}
\vspace{-10pt}
\end{figure*}

\begin{figure*}[t!]
\centering
\includegraphics[width=1.0\textwidth]{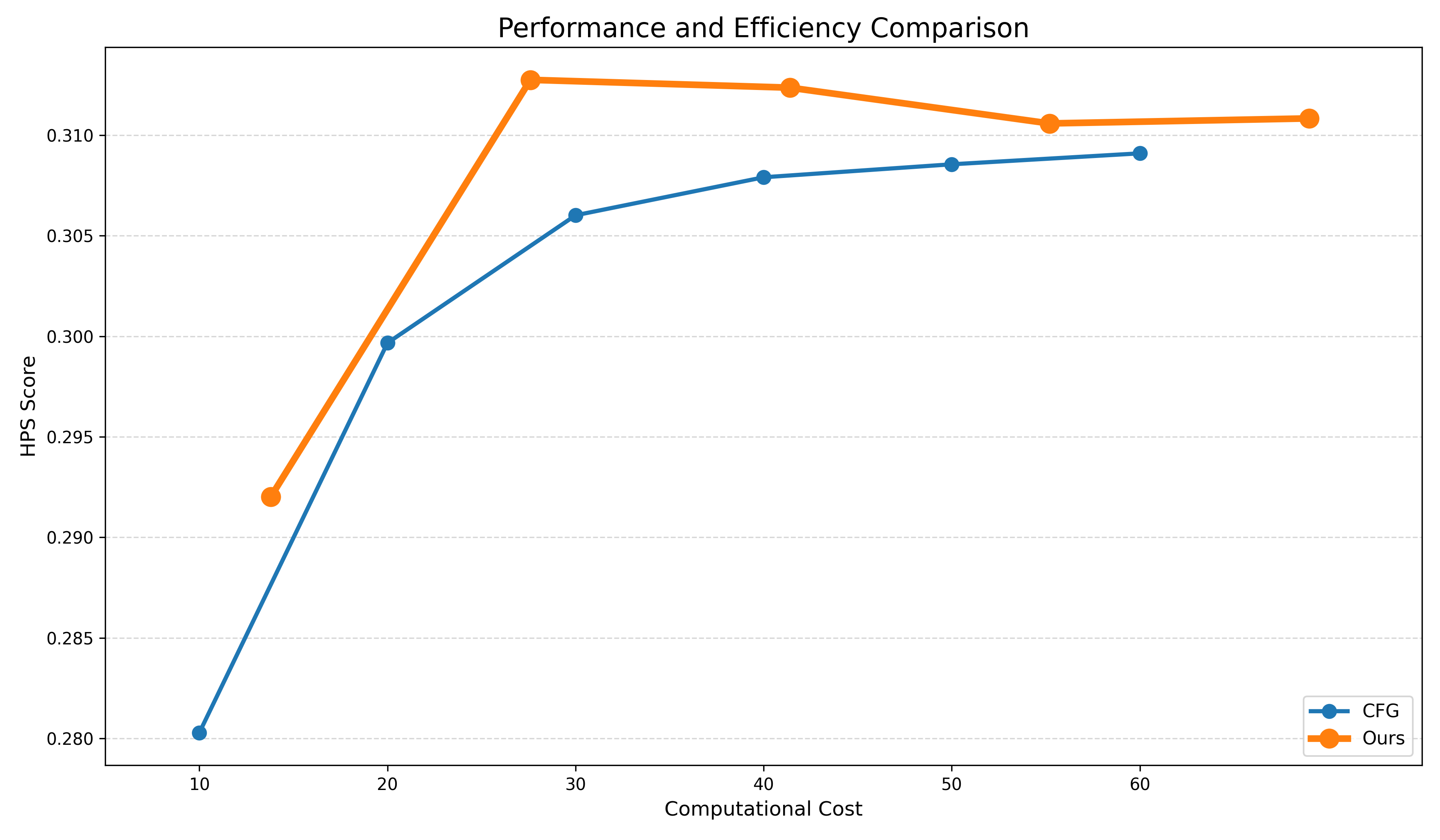}
\caption{
    \textbf{Performance-Efficiency Trade-off Analysis.}
    This figure compares our method against CFG by plotting HPS Score as a function of computational cost. 
    (Curves positioned further toward the \textbf{top-left} indicate superior methods.)
    The x-axis represents a normalized computational cost, where the cost for CFG equals its inference steps, while the cost for our method is scaled by a factor of 1.4 to reflect its $\sim$40\% computational overhead. 
    The plot illustrates that our method offers a significantly better trade-off. 
    For instance, our method with just 20 inference steps (equivalent cost $\approx$ 28) already achieves a higher HPS score than CFG at 60 steps. 
    This demonstrates that our method yields substantial quality improvements for a comparable or even lower computational budget.
}
\label{fig:hps_vs_steps}
\vspace{-5pt}
\end{figure*}

\begin{figure*}[t!]
\centering
\includegraphics[width=1.0\textwidth]{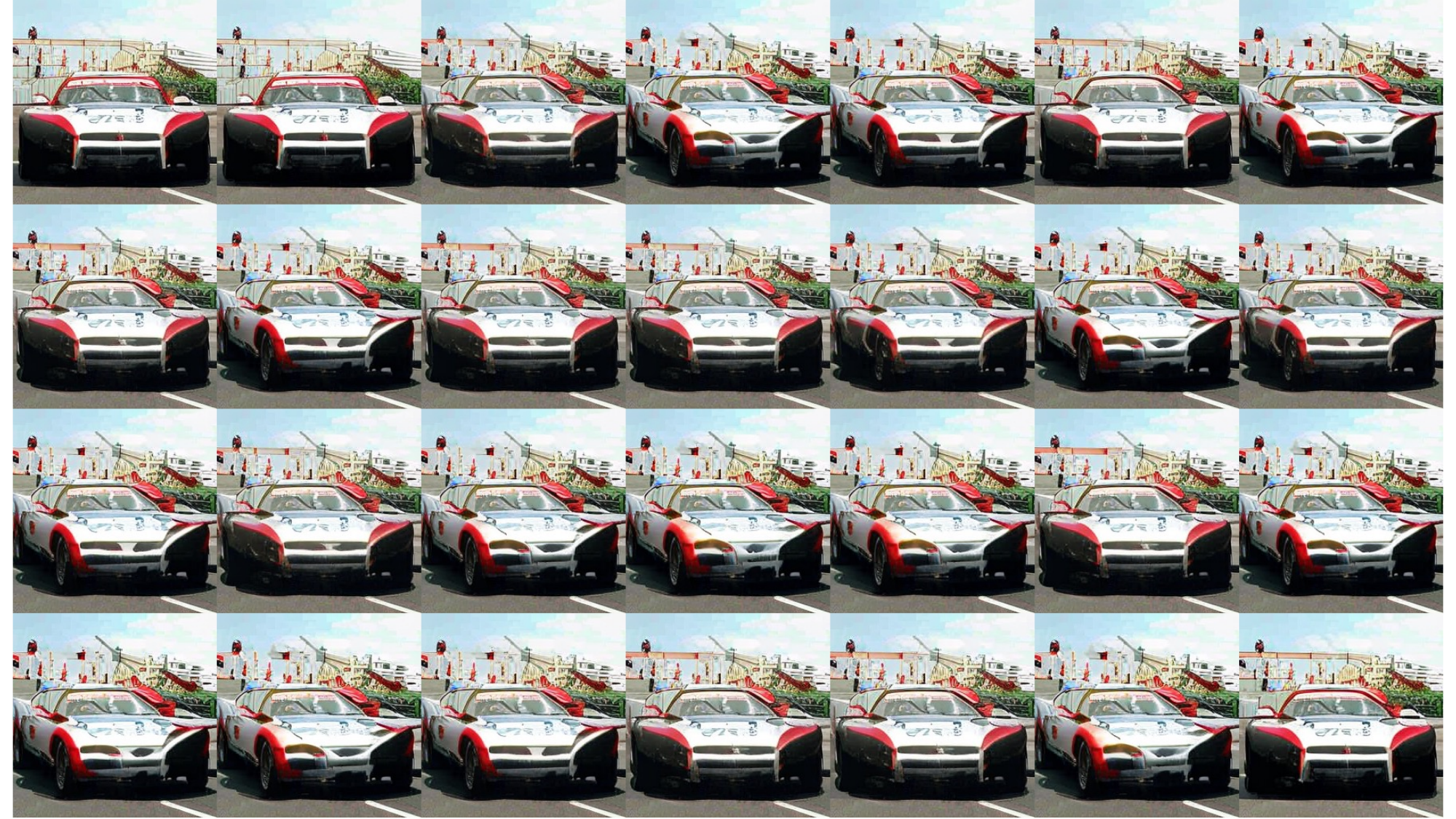}
\caption{
    \textbf{Impact of dropping a single, fixed transformer block in SiT-XL.} 
    Each of the 28 images corresponds to dropping one specific block for all timesteps on the ImageNet 256$\times$256 task. 
    The visual consistency across the grid demonstrates the model's robustness against block-level perturbations.
}
\label{fig:vis_drop_block}
\vspace{-5pt}
\end{figure*}

\section{Future Work}
Our method is training-free and plug-and-play, requiring no additional fine-tuning or retraining of the underlying model. This enables seamless integration into existing pipelines and allows practitioners to improve generation quality immediately at inference time.
A promising direction is to further explore and generalize the proposed self-guidance mechanism, which is designed to mitigate uncertainty regions within the model and thereby correct suboptimal trajectories. This capability suggests broad applicability to tasks where small deviations can lead to noticeable quality degradation. In the image domain, more reliable internal guidance may improve (i) image or video editing \citep{zhu2024instantswap, wang2024cove, wang2024taming} accuracy, where faithful execution of edit instructions is critical; (ii) human-preference alignment objectives \citep{t2irl, ma2024followyouremoji, Huang_2025_ICCV}; and (iii) high-fidelity image synthesis \citep{epg, chen2025storyctrl} under challenging prompts. In the video domain, the same principle may translate to improved motion coherence and temporal smoothness, potentially reducing temporal artifacts \citep{wangprecisecache, omni, su2025safe, wu2025imagerysearch, ling2025vmbench,feng2025narrlv, zhu2026artifact, ma2025follow}.
Finally, our observations are grounded in the redundancy of transformer architectures, suggesting potential applicability beyond DiT architectures. An important future direction is to investigate whether analogous self-guidance mechanisms can improve faithfulness, robustness in LLMs and MLLMs, as well as enable broader potential applications \citep{yao2025understanding, yu2025physicsminions, xie2024uncertainty, wang2025fast, wang2026exposing, wei2023efficient, feng2025seeing, fang2024real, fangphoton, jiang2025world4rl, xie2026q, huang2025diffusiondatasetcondensationtraining, chu2025gpg, chu2021twins, chu2021conditional, huang2025taming}.

%% file: iclr2026_conference.bib
@article{croitoru2023diffusion,
  title={Diffusion models in vision: A survey},
  author={Croitoru, Florinel-Alin and Hondru, Vlad and Ionescu, Radu Tudor and Shah, Mubarak},
  journal={IEEE transactions on pattern analysis and machine intelligence},
  volume={45},
  number={9},
  pages={10850--10869},
  year={2023},
  publisher={Ieee}
}

@misc{wu2025qwenimagetechnicalreport,
      title={Qwen-Image Technical Report}, 
      author={Chenfei Wu and Jiahao Li and Jingren Zhou and Junyang Lin and Kaiyuan Gao and Kun Yan and Sheng-ming Yin and Shuai Bai and Xiao Xu and Yilei Chen and Yuxiang Chen and Zecheng Tang and Zekai Zhang and Zhengyi Wang and An Yang and Bowen Yu and Chen Cheng and Dayiheng Liu and Deqing Li and Hang Zhang and Hao Meng and Hu Wei and Jingyuan Ni and Kai Chen and Kuan Cao and Liang Peng and Lin Qu and Minggang Wu and Peng Wang and Shuting Yu and Tingkun Wen and Wensen Feng and Xiaoxiao Xu and Yi Wang and Yichang Zhang and Yongqiang Zhu and Yujia Wu and Yuxuan Cai and Zenan Liu},
      year={2025},
      eprint={2508.02324},
      archivePrefix={arXiv},
      primaryClass={cs.CV},
      url={https://arxiv.org/abs/2508.02324}, 
}

@article{sde,
  title={Score-based generative modeling through stochastic differential equations},
  author={Song, Yang and Sohl-Dickstein, Jascha and Kingma, Diederik P and Kumar, Abhishek and Ermon, Stefano and Poole, Ben},
  journal={arXiv preprint arXiv:2011.13456},
  year={2020}
}

@article{diffusionbeatsgans,
  title={Diffusion models beat gans on image synthesis},
  author={Dhariwal, Prafulla and Nichol, Alexander},
  journal={Advances in neural information processing systems},
  volume={34},
  pages={8780--8794},
  year={2021}
}

@article{t2irl,
  title={Taming Preference Mode Collapse via Directional Decoupling Alignment in Diffusion Reinforcement Learning},
  author={Chen, Chubin and Hu, Sujie and Zhu, Jiashu and Wu, Meiqi and Chen, Jintao and Li, Yanxun and Huang, Nisha and Fang, Chengyu and Wu, Jiahong and Chu, Xiangxiang and others},
  journal={arXiv preprint arXiv:2512.24146},
  year={2025}
}

@article{cfg,
  title={Classifier-free diffusion guidance},
  author={Ho, Jonathan and Salimans, Tim},
  journal={arXiv preprint arXiv:2207.12598},
  year={2022}
}

@inproceedings{apg,
  title={Eliminating oversaturation and artifacts of high guidance scales in diffusion models},
  author={Sadat, Seyedmorteza and Hilliges, Otmar and Weber, Romann M},
  booktitle={The Thirteenth International Conference on Learning Representations},
  year={2024}
}

@article{cfg++,
  title={Cfg++: Manifold-constrained classifier free guidance for diffusion models},
  author={Chung, Hyungjin and Kim, Jeongsol and Park, Geon Yeong and Nam, Hyelin and Ye, Jong Chul},
  journal={arXiv preprint arXiv:2406.08070},
  year={2024}
}

@article{intervalcfg,
  title={Applying guidance in a limited interval improves sample and distribution quality in diffusion models},
  author={Kynk{\"a}{\"a}nniemi, Tuomas and Aittala, Miika and Karras, Tero and Laine, Samuli and Aila, Timo and Lehtinen, Jaakko},
  journal={Advances in Neural Information Processing Systems},
  volume={37},
  pages={122458--122483},
  year={2024}
}

@article{cfgzero,
  title={Cfg-zero*: Improved classifier-free guidance for flow matching models},
  author={Fan, Weichen and Zheng, Amber Yijia and Yeh, Raymond A and Liu, Ziwei},
  journal={arXiv preprint arXiv:2503.18886},
  year={2025}
}

@article{adg,
  title={Angle Domain Guidance: Latent Diffusion Requires Rotation Rather Than Extrapolation},
  author={Jin, Cheng and Xiao, Zhenyu and Liu, Chutao and Gu, Yuantao},
  journal={arXiv preprint arXiv:2506.11039},
  year={2025}
}

@article{autoguidance,
  title={Guiding a diffusion model with a bad version of itself},
  author={Karras, Tero and Aittala, Miika and Kynk{\"a}{\"a}nniemi, Tuomas and Lehtinen, Jaakko and Aila, Timo and Laine, Samuli},
  journal={Advances in Neural Information Processing Systems},
  volume={37},
  pages={52996--53021},
  year={2024}
}

@inproceedings{sag,
  title={Improving sample quality of diffusion models using self-attention guidance},
  author={Hong, Susung and Lee, Gyuseong and Jang, Wooseok and Kim, Seungryong},
  booktitle={Proceedings of the IEEE/CVF International Conference on Computer Vision},
  pages={7462--7471},
  year={2023}
}

@article{seg,
  title={Smoothed energy guidance: Guiding diffusion models with reduced energy curvature of attention},
  author={Hong, Susung},
  journal={Advances in Neural Information Processing Systems},
  volume={37},
  pages={66743--66772},
  year={2024}
}

@inproceedings{pag,
  title={Self-rectifying diffusion sampling with perturbed-attention guidance},
  author={Ahn, Donghoon and Cho, Hyoungwon and Min, Jaewon and Jang, Wooseok and Kim, Jungwoo and Kim, SeonHwa and Park, Hyun Hee and Jin, Kyong Hwan and Kim, Seungryong},
  booktitle={European Conference on Computer Vision},
  pages={1--17},
  year={2024},
  organization={Springer}
}

@inproceedings{stg,
  title={Spatiotemporal skip guidance for enhanced video diffusion sampling},
  author={Hyung, Junha and Kim, Kinam and Hong, Susung and Kim, Min-Jung and Choo, Jaegul},
  booktitle={Proceedings of the Computer Vision and Pattern Recognition Conference},
  pages={11006--11015},
  year={2025}
}

@article{spg,
  title={SPG: Improving Motion Diffusion by Smooth Perturbation Guidance},
  author={Jeon, Boseong},
  journal={arXiv preprint arXiv:2503.02577},
  year={2025}
}

@article{gauss1,
  title={Verifying the union of manifolds hypothesis for image data},
  author={Brown, Bradley CA and Caterini, Anthony L and Ross, Brendan Leigh and Cresswell, Jesse C and Loaiza-Ganem, Gabriel},
  journal={arXiv preprint arXiv:2207.02862},
  year={2022}
}

@inproceedings{gauss2,
  author    = {Phillip Pope and
               Chen Zhu and
               Ahmed Abdelkader and
               Micah Goldblum and
               Tom Goldstein},
  title     = {The Intrinsic Dimension of Images and Its Impact on Learning},
  booktitle = {9th International Conference on Learning Representations, {ICLR} 2021,
               Virtual Event, Austria, May 3-7, 2021},
  publisher = {OpenReview.net},
  year      = {2021},
  url       = {https://openreview.net/forum?id=XJk19XzGq2J},
  bibsource = {dblp computer science bibliography, https://dblp.org}
}

@article{redundant1,
  title={Token caching for diffusion transformer acceleration},
  author={Lou, Jinming and Luo, Wenyang and Liu, Yufan and Li, Bing and Ding, Xinmiao and Hu, Weiming and Cao, Jiajiong and Li, Yuming and Ma, Chenguang},
  journal={arXiv preprint arXiv:2409.18523},
  year={2024}
}

@inproceedings{redudant2_stableflow,
  title={Stable flow: Vital layers for training-free image editing},
  author={Avrahami, Omri and Patashnik, Or and Fried, Ohad and Nemchinov, Egor and Aberman, Kfir and Lischinski, Dani and Cohen-Or, Daniel},
  booktitle={Proceedings of the Computer Vision and Pattern Recognition Conference},
  pages={7877--7888},
  year={2025}
}

@article{redundant3,
  title={Ditfastattn: Attention compression for diffusion transformer models},
  author={Yuan, Zhihang and Zhang, Hanling and Pu, Lu and Ning, Xuefei and Zhang, Linfeng and Zhao, Tianchen and Yan, Shengen and Dai, Guohao and Wang, Yu},
  journal={Advances in Neural Information Processing Systems},
  volume={37},
  pages={1196--1219},
  year={2024}
}

@inproceedings{dropout,
  title={Dropout as a bayesian approximation: Representing model uncertainty in deep learning},
  author={Gal, Yarin and Ghahramani, Zoubin},
  booktitle={international conference on machine learning},
  pages={1050--1059},
  year={2016},
  organization={PMLR}
}

@article{song2020denoising,
  title={Denoising diffusion implicit models},
  author={Song, Jiaming and Meng, Chenlin and Ermon, Stefano},
  journal={arXiv preprint arXiv:2010.02502},
  year={2020}
}

@article{ho2020denoising,
  title={Denoising diffusion probabilistic models},
  author={Ho, Jonathan and Jain, Ajay and Abbeel, Pieter},
  journal={Advances in neural information processing systems},
  volume={33},
  pages={6840--6851},
  year={2020}
}

@inproceedings{rombach2022high,
  title={High-resolution image synthesis with latent diffusion models},
  author={Rombach, Robin and Blattmann, Andreas and Lorenz, Dominik and Esser, Patrick and Ommer, Bj{\"o}rn},
  booktitle={Proceedings of the IEEE/CVF conference on computer vision and pattern recognition},
  pages={10684--10695},
  year={2022}
}

@article{podell2023sdxl,
  title={Sdxl: Improving latent diffusion models for high-resolution image synthesis},
  author={Podell, Dustin and English, Zion and Lacey, Kyle and Blattmann, Andreas and Dockhorn, Tim and M{\"u}ller, Jonas and Penna, Joe and Rombach, Robin},
  journal={arXiv preprint arXiv:2307.01952},
  year={2023}
}

@article{wan2025wan,
  title={Wan: Open and advanced large-scale video generative models},
  author={Wan, Team and Wang, Ang and Ai, Baole and Wen, Bin and Mao, Chaojie and Xie, Chen-Wei and Chen, Di and Yu, Feiwu and Zhao, Haiming and Yang, Jianxiao and others},
  journal={arXiv preprint arXiv:2503.20314},
  year={2025}
}

@article{kong2024hunyuanvideo,
  title={Hunyuanvideo: A systematic framework for large video generative models},
  author={Kong, Weijie and Tian, Qi and Zhang, Zijian and Min, Rox and Dai, Zuozhuo and Zhou, Jin and Xiong, Jiangfeng and Li, Xin and Wu, Bo and Zhang, Jianwei and others},
  journal={arXiv preprint arXiv:2412.03603},
  year={2024}
}

@article{bradley2024classifier,
  title={Classifier-free guidance is a predictor-corrector},
  author={Bradley, Arwen and Nakkiran, Preetum},
  journal={arXiv preprint arXiv:2408.09000},
  year={2024}
}

@inproceedings{peebles2023scalable,
  title={Scalable diffusion models with transformers},
  author={Peebles, William and Xie, Saining},
  booktitle={Proceedings of the IEEE/CVF international conference on computer vision},
  pages={4195--4205},
  year={2023}
}

@inproceedings{
chen2024revealing,
title={Revealing the Dark Secrets of Extremely Large Kernel ConvNets on Robustness},
author={Honghao Chen and Yurong Zhang and Xiaokun Feng and Xiangxiang Chu and Kaiqi Huang},
booktitle={Forty-first International Conference on Machine Learning},
year={2024},
url={https://openreview.net/forum?id=rkYOxLLv2x}
}

@article{wu2023human,
  title={Human preference score v2: A solid benchmark for evaluating human preferences of text-to-image synthesis},
  author={Wu, Xiaoshi and Hao, Yiming and Sun, Keqiang and Chen, Yixiong and Zhu, Feng and Zhao, Rui and Li, Hongsheng},
  journal={arXiv preprint arXiv:2306.09341},
  year={2023}
}

@article{huang2023t2i,
  title={T2i-compbench: A comprehensive benchmark for open-world compositional text-to-image generation},
  author={Huang, Kaiyi and Sun, Kaiyue and Xie, Enze and Li, Zhenguo and Liu, Xihui},
  journal={Advances in Neural Information Processing Systems},
  volume={36},
  pages={78723--78747},
  year={2023}
}

@article{wu2023q,
  title={Q-align: Teaching lmms for visual scoring via discrete text-defined levels},
  author={Wu, Haoning and Zhang, Zicheng and Zhang, Weixia and Chen, Chaofeng and Liao, Liang and Li, Chunyi and Gao, Yixuan and Wang, Annan and Zhang, Erli and Sun, Wenxiu and others},
  journal={arXiv preprint arXiv:2312.17090},
  year={2023}
}

@inproceedings{huang2024vbench,
  title={Vbench: Comprehensive benchmark suite for video generative models},
  author={Huang, Ziqi and He, Yinan and Yu, Jiashuo and Zhang, Fan and Si, Chenyang and Jiang, Yuming and Zhang, Yuanhan and Wu, Tianxing and Jin, Qingyang and Chanpaisit, Nattapol and others},
  booktitle={Proceedings of the IEEE/CVF Conference on Computer Vision and Pattern Recognition},
  pages={21807--21818},
  year={2024}
}

@misc{sd3.5,
  title={Stable diffusion 3.5},
  author={AI, S},
  year={2024}
}

@inproceedings{sd3,
  title={Scaling rectified flow transformers for high-resolution image synthesis},
  author={Esser, Patrick and Kulal, Sumith and Blattmann, Andreas and Entezari, Rahim and M{\"u}ller, Jonas and Saini, Harry and Levi, Yam and Lorenz, Dominik and Sauer, Axel and Boesel, Frederic and others},
  booktitle={Forty-first international conference on machine learning},
  year={2024}
}

@inproceedings{ling2025vmbench,
  title={VMBench: A Benchmark for Perception-Aligned Video Motion Generation},
  author={Ling, Xinran and Zhu, Chen and Wu, Meiqi and Li, Hangyu and Feng, Xiaokun and Yang, Cundian and Hao, Aiming and Zhu, Jiashu and Wu, Jiahong and Chu, Xiangxiang},
  booktitle={ICCV},
  year={2025}
}

@article{feng2025narrlv,
  title={NarrLV: Towards a Comprehensive Narrative-Centric Evaluation for Long Video Generation Models},
  author={Feng, Xiaokun and Yu, Haiming and Wu, Meiqi and Hu, Shiyu and Chen, Jintao and Zhu, Chen and Wu, Jiahong and Chu, Xiangxiang and Huang, Kaiqi},
  journal={arXiv preprint arXiv:2507.11245},
  year={2025}
}

@article{liu2024survey,
  title={A survey of ai-generated video evaluation},
  author={Liu, Xiao and Xiang, Xinhao and Li, Zizhong and Wang, Yongheng and Li, Zhuoheng and Liu, Zhuosheng and Zhang, Weidi and Ye, Weiqi and Zhang, Jiawei},
  journal={arXiv preprint arXiv:2410.19884},
  year={2024}
}

@misc{flux,
    author={Black Forest Labs},
    title={FLUX},
    year={2024},
    howpublished={\url{https://github.com/black-forest-labs/flux}},
}

@inproceedings{chen2025finger,
  title={FingER: Content Aware Fine-grained Evaluation with Reasoning for AI-Generated Videos},
  author={Chen, Rui and Sun, Lei and Tang, Jing and Li, Geng and Chu, Xiangxiang},
  booktitle={ACM MM},
  year={2025}
}

@inproceedings{chu2025usp,
  title={Usp: Unified self-supervised pretraining for image generation and understanding},
  author={Chu, Xiangxiang and Li, Renda and Wang, Yong},
  booktitle={ICCV},
  year={2025}
}

@article{fang2025integrating,
  title={Integrating extra modality helps segmentor find camouflaged objects well},
  author={Fang, Chengyu and He, Chunming and Tang, Longxiang and Zhang, Yuelin and Zhu, Chenyang and Shen, Yuqi and Chen, Chubin and Xu, Guoxia and Li, Xiu},
  journal={arXiv preprint arXiv:2502.14471},
  year={2025}
}

@article{he2025diffusion,
  title={Diffusion models in low-level vision: A survey},
  author={He, Chunming and Shen, Yuqi and Fang, Chengyu and Xiao, Fengyang and Tang, Longxiang and Zhang, Yulun and Zuo, Wangmeng and Guo, Zhenhua and Li, Xiu},
  journal={IEEE Transactions on Pattern Analysis and Machine Intelligence},
  year={2025},
  publisher={IEEE}
}

@article{wang2024cove,
  title={Cove: Unleashing the diffusion feature correspondence for consistent video editing},
  author={Wang, Jiangshan and Ma, Yue and Guo, Jiayi and Xiao, Yicheng and Huang, Gao and Li, Xiu},
  journal={arXiv preprint arXiv:2406.08850},
  year={2024}
}

@article{wang2024taming,
  title={Taming rectified flow for inversion and editing},
  author={Wang, Jiangshan and Pu, Junfu and Qi, Zhongang and Guo, Jiayi and Ma, Yue and Huang, Nisha and Chen, Yuxin and Li, Xiu and Shan, Ying},
  journal={arXiv preprint arXiv:2411.04746},
  year={2024}
}

@article{huang2023region,
title={Region-aware diffusion for zero-shot text-driven image editing},
author={Huang, Nisha and Tang, Fan and Dong, Weiming and Lee, Tong-Yee and Xu, Changsheng},
journal={arXiv preprint arXiv:2302.11797},
year={2023}
}

@article{huang2024diffstyler,
title={Diffstyler: Controllable dual diffusion for text-driven image stylization},
author={Huang, Nisha and Zhang, Yuxin and Tang, Fan and Ma, Chongyang and Huang, Haibin and Dong, Weiming and Xu, Changsheng},
journal={IEEE Transactions on Neural Networks and Learning Systems},
year={2024},
publisher={IEEE}
}

@article{zhu2024instantswap,
  title={InstantSwap: Fast Customized Concept Swapping across Sharp Shape Differences},
  author={Zhu, Chenyang and Li, Kai and Ma, Yue and Tang, Longxiang and Fang, Chengyu and Chen, Chubin and Chen, Qifeng and Li, Xiu},
  journal={arXiv preprint arXiv:2412.01197},
  year={2024}
}

@inproceedings{chu2024visionllama,
  title={Visionllama: A unified llama backbone for vision tasks},
  author={Chu, Xiangxiang and Su, Jianlin and Zhang, Bo and Shen, Chunhua},
  booktitle={European Conference on Computer Vision},
  pages={1--18},
  year={2024},
  organization={Springer}
}

@misc{flow1,
      title={Flow Matching for Generative Modeling}, 
      author={Yaron Lipman and Ricky T. Q. Chen and Heli Ben-Hamu and Maximilian Nickel and Matt Le},
      year={2023},
      eprint={2210.02747},
      archivePrefix={arXiv},
      primaryClass={cs.LG},
      url={https://arxiv.org/abs/2210.02747}, 
}

@misc{flow2,
      title={Flow Straight and Fast: Learning to Generate and Transfer Data with Rectified Flow}, 
      author={Xingchao Liu and Chengyue Gong and Qiang Liu},
      year={2022},
      eprint={2209.03003},
      archivePrefix={arXiv},
      primaryClass={cs.LG},
      url={https://arxiv.org/abs/2209.03003}, 
}

@article{omni,
  title={Omni-effects: Unified and spatially-controllable visual effects generation},
  author={Mao, Fangyuan and Hao, Aiming and Chen, Jintao and Liu, Dongxia and Feng, Xiaokun and Zhu, Jiashu and Wu, Meiqi and Chen, Chubin and Wu, Jiahong and Chu, Xiangxiang},
  journal={arXiv preprint arXiv:2508.07981},
  year={2025}
}

@misc{flow3,
      title={Discrete Flow Matching}, 
      author={Itai Gat and Tal Remez and Neta Shaul and Felix Kreuk and Ricky T. Q. Chen and Gabriel Synnaeve and Yossi Adi and Yaron Lipman},
      year={2024},
      eprint={2407.15595},
      archivePrefix={arXiv},
      primaryClass={cs.LG},
      url={https://arxiv.org/abs/2407.15595}, 
}

@misc{flow5,
      title={Convergence Analysis for General Probability Flow ODEs of Diffusion Models in Wasserstein Distances}, 
      author={Xuefeng Gao and Lingjiong Zhu},
      year={2025},
      eprint={2401.17958},
      archivePrefix={arXiv},
      primaryClass={stat.ML},
      url={https://arxiv.org/abs/2401.17958}, 
}

@inproceedings{ma2024followpose,
  title={Follow your pose: Pose-guided text-to-video generation using pose-free videos},
  author={Ma, Yue and He, Yingqing and Cun, Xiaodong and Wang, Xintao and Chen, Siran and Li, Xiu and Chen, Qifeng},
  booktitle={Proceedings of the AAAI Conference on Artificial Intelligence},
  volume={38},
  number={5},
  pages={4117--4125},
  year={2024}
}

@inproceedings{ma2025followyourclick,
  title={Follow-Your-Click: Open-domain Regional Image Animation via Motion Prompts},
  author={Ma, Yue and He, Yingqing and Wang, Hongfa and Wang, Andong and Shen, Leqi and Qi, Chenyang and Ying, Jixuan and Cai, Chengfei and Li, Zhifeng and Shum, Heung-Yeung and others},
  booktitle={Proceedings of the AAAI Conference on Artificial Intelligence},
  volume={39},
  number={6},
  pages={6018--6026},
  year={2025}
}

@inproceedings{ma2024followyouremoji,
  title={Follow-your-emoji: Fine-controllable and expressive freestyle portrait animation},
  author={Ma, Yue and Liu, Hongyu and Wang, Hongfa and Pan, Heng and He, Yingqing and Yuan, Junkun and Zeng, Ailing and Cai, Chengfei and Shum, Heung-Yeung and Liu, Wei and others},
  booktitle={SIGGRAPH Asia 2024 Conference Papers},
  pages={1--12},
  year={2024}
}

@misc{attn1,
      title={FateZero: Fusing Attentions for Zero-shot Text-based Video Editing}, 
      author={Chenyang Qi and Xiaodong Cun and Yong Zhang and Chenyang Lei and Xintao Wang and Ying Shan and Qifeng Chen},
      year={2023},
      eprint={2303.09535},
      archivePrefix={arXiv},
      primaryClass={cs.CV},
      url={https://arxiv.org/abs/2303.09535}, 
}

@misc{attn2,
      title={LIME: Localized Image Editing via Attention Regularization in Diffusion Models}, 
      author={Enis Simsar and Alessio Tonioni and Yongqin Xian and Thomas Hofmann and Federico Tombari},
      year={2024},
      eprint={2312.09256},
      archivePrefix={arXiv},
      primaryClass={cs.CV},
      url={https://arxiv.org/abs/2312.09256}, 
}

@misc{sit,
      title={SiT: Exploring Flow and Diffusion-based Generative Models with Scalable Interpolant Transformers}, 
      author={Nanye Ma and Mark Goldstein and Michael S. Albergo and Nicholas M. Boffi and Eric Vanden-Eijnden and Saining Xie},
      year={2024},
      eprint={2401.08740},
      archivePrefix={arXiv},
      primaryClass={cs.CV},
      url={https://arxiv.org/abs/2401.08740}, 
}

@article{deepensembles,
  title={Simple and scalable predictive uncertainty estimation using deep ensembles},
  author={Lakshminarayanan, Balaji and Pritzel, Alexander and Blundell, Charles},
  journal={Advances in neural information processing systems},
  volume={30},
  year={2017}
}

@article{snapshot,
  title={Snapshot ensembles: Train 1, get m for free},
  author={Huang, Gao and Li, Yixuan and Pleiss, Geoff and Liu, Zhuang and Hopcroft, John E and Weinberger, Kilian Q},
  journal={arXiv preprint arXiv:1704.00109},
  year={2017}
}

@article{batchensemble,
  title={Batchensemble: an alternative approach to efficient ensemble and lifelong learning},
  author={Wen, Yeming and Tran, Dustin and Ba, Jimmy},
  journal={arXiv preprint arXiv:2002.06715},
  year={2020}
}

@article{epg,
  title={There is no vae: End-to-end pixel-space generative modeling via self-supervised pre-training},
  author={Lei, Jiachen and Liu, Keli and Berner, Julius and Yu, Haiming and Zheng, Hongkai and Wu, Jiahong and Chu, Xiangxiang},
  journal={arXiv preprint arXiv:2510.12586},
  year={2025}
}

@inproceedings{wangprecisecache,
  title={PreciseCache: Precise Feature Caching for Efficient and High-fidelity Video Generation},
  author={Wang, Jiangshan and Zhao, Kang and Guo, Jiayi and Wang, Jiayu and Guo, Hang and Zhu, Chenyang and Yue, Xiangyu and Li, Xiu},
  booktitle={The Fourteenth International Conference on Learning Representations}
}

@misc{
chen2025storyctrl,
title={StoryCtrl: Customized Story Visualization with Fine-Grained Control},
author={Chubin Chen and Jiashu Zhu and Chenyang Zhu and Jiangshan Wang and Nisha Huang and Chengyu Fang and Jiahong Wu and Xiangxiang Chu and Xiu Li},
year={2025},
url={https://openreview.net/forum?id=hQv6a2sSwo}
}

@article{lei2023masked,
  title={Masked diffusion models are fast distribution learners},
  author={Lei, Jiachen and Wang, Qinglong and Cheng, Peng and Ba, Zhongjie and Qin, Zhan and Wang, Zhibo and Liu, Zhenguang and Ren, Kui},
  journal={arXiv preprint arXiv:2306.11363},
  year={2023}
}

@article{su2025safe,
  title={Safe-Sora: Safe Text-to-Video Generation via Graphical Watermarking},
  author={Su, Zihan and Qiu, Xuerui and Xu, Hongbin and Jiang, Tangyu and Zhuang, Junhao and Yuan, Chun and Li, Ming and He, Shengfeng and Yu, Fei Richard},
  journal={arXiv preprint arXiv:2505.12667},
  year={2025}
}

@article{wu2025imagerysearch,
  title={ImagerySearch: Adaptive Test-Time Search for Video Generation Beyond Semantic Dependency Constraints},
  author={Wu, Meiqi and Zhu, Jiashu and Feng, Xiaokun and Chen, Chubin and Zhu, Chen and Song, Bingze and Mao, Fangyuan and Wu, Jiahong and Chu, Xiangxiang and Huang, Kaiqi},
  journal={arXiv preprint arXiv:2510.14847},
  year={2025}
}

@InProceedings{Huang_2025_ICCV,
    author    = {Huang, Nisha and Liu, Henglin and Lin, Yizhou and Huang, Kaer and Chen, Chubin and Guo, Jie and Lee, Tong-yee and Li, Xiu},
    title     = {MaTe: Images Are All You Need for Material Transfer via Diffusion Transformer},
    booktitle = {Proceedings of the IEEE/CVF International Conference on Computer Vision (ICCV)},
    month     = {October},
    year      = {2025},
    pages     = {15117-15126}
}

@article{xie2026q,
  title={Q-Hawkeye: Reliable Visual Policy Optimization for Image Quality Assessment},
  author={Xie, Wulin and Dai, Rui and Ding, Ruidong and Liu, Kaikui and Chu, Xiangxiang and Hou, Xinwen and Wen, Jie},
  journal={arXiv preprint arXiv:2601.22920},
  year={2026}
}

@inproceedings{xie2024uncertainty,
  title={Uncertainty-aware pseudo-labeling and dual graph driven network for incomplete multi-view multi-label classification},
  author={Xie, Wulin and Lu, Xiaohuan and Liu, Yadong and Long, Jiang and Zhang, Bob and Zhao, Shuping and Wen, Jie},
  booktitle={Proceedings of the 32nd ACM International Conference on Multimedia},
  pages={6656--6665},
  year={2024}
}

@article{yao2025understanding,
  title={Understanding the repeat curse in large language models from a feature perspective},
  author={Yao, Junchi and Yang, Shu and Xu, Jianhua and Hu, Lijie and Li, Mengdi and Wang, Di},
  journal={arXiv preprint arXiv:2504.14218},
  year={2025}
}

@article{yu2025physicsminions,
  title={Physicsminions: Winning gold medals in the latest physics olympiads with a coevolutionary multimodal multi-agent system},
  author={Yu, Fangchen and Yao, Junchi and Wang, Ziyi and Wan, Haiyuan and Huang, Youling and Zhang, Bo and Hu, Shuyue and Zhou, Dongzhan and Ding, Ning and Cui, Ganqu and others},
  journal={arXiv preprint arXiv:2509.24855},
  year={2025}
}

@article{wang2026exposing,
  title={Exposing and Defending the Achilles' Heel of Video Mixture-of-Experts},
  author={Wang, Songping and Liu, Qinglong and Lyu, Yueming and Li, Ning and He, Ziwen and Shan, Caifeng},
  journal={arXiv preprint arXiv:2602.01369},
  year={2026}
}

@article{wei2023efficient,
  title={Efficient robustness assessment via adversarial spatial-temporal focus on videos},
  author={Wei, Xingxing and Wang, Songping and Yan, Huanqian},
  journal={IEEE Transactions on Pattern Analysis and Machine Intelligence},
  volume={45},
  number={9},
  pages={10898--10912},
  year={2023},
  publisher={IEEE}
}

@article{wang2025fast,
  title={Fast adversarial training with weak-to-strong spatial-temporal consistency in the frequency domain on videos},
  author={Wang, Songping and Liu, Hanqing and Lyu, Yueming and Hu, Xiantao and He, Ziwen and Wang, Wei and Shan, Caifeng and Wang, Liang},
  journal={IEEE Transactions on Information Forensics and Security},
  volume={21},
  pages={681--696},
  year={2025},
  publisher={IEEE}
}

@article{feng2025seeing,
  title={Seeing across views: Benchmarking spatial reasoning of vision-language models in robotic scenes},
  author={Feng, Zhiyuan and Kang, Zhaolu and Wang, Qijie and Du, Zhiying and Yan, Jiongrui and Shi, Shubin and Yuan, Chengbo and Liang, Huizhi and Deng, Yu and Li, Qixiu and others},
  journal={arXiv preprint arXiv:2510.19400},
  year={2025}
}

@inproceedings{fangphoton,
  title={Photon: Speedup Volume Understanding with Efficient Multimodal Large Language Models},
  author={Fang, Chengyu and Guo, Heng and Jiang, Zheng and He, Chunming and Li, Xiu and Xu, Minfeng},
  booktitle={The Fourteenth International Conference on Learning Representations},
  year={2026}
}

@inproceedings{fang2024real,
 title={Real-world Image Dehazing with Coherence-based Pseudo Labeling and Cooperative Unfolding Network},
 author={Fang, Chengyu and He, Chunming and Xiao, Fengyang and Zhang, Yulun and Tang, Longxiang and Zhang, Yuelin and Li, Kai and Li, Xiu},
 booktitle={The Thirty-eighth Annual Conference on Neural Information Processing Systems},
 year={2024}
}

@article{jiang2025world4rl,
  title={World4rl: Diffusion world models for policy refinement with reinforcement learning for robotic manipulation},
  author={Jiang, Zhennan and Liu, Kai and Qin, Yuxin and Tian, Shuai and Zheng, Yupeng and Zhou, Mingcai and Yu, Chao and Li, Haoran and Zhao, Dongbin},
  journal={arXiv preprint arXiv:2509.19080},
  year={2025}
}

@misc{huang2025diffusiondatasetcondensationtraining,
      title={Diffusion Dataset Condensation: Training Your Diffusion Model Faster with Less Data}, 
      author={Rui Huang and Shitong Shao and Zikai Zhou and Pukun Zhao and Hangyu Guo and Tian Ye and Lichen Bai and Shuo Yang and Zeke Xie},
      year={2025},
      eprint={2507.05914},
      archivePrefix={arXiv},
      primaryClass={cs.LG},
      url={https://arxiv.org/abs/2507.05914}, 
}

@article{chu2025gpg,
  title={Gpg: A simple and strong reinforcement learning baseline for model reasoning},
  author={Chu, Xiangxiang and Huang, Hailang and Zhang, Xiao and Wei, Fei and Wang, Yong},
  journal={arXiv preprint arXiv:2504.02546},
  year={2025}
}

@article{zhu2026artifact,
  title={Artifact-Aware Evaluation for High-Quality Video Generation},
  author={Zhu, Chen and Zhu, Jiashu and Li, Yanxun and Wu, Meiqi and Song, Bingze and Chen, Chubin and Wu, Jiahong and Chu, Xiangxiang and Wang, Yangang},
  journal={arXiv preprint arXiv:2601.20297},
  year={2026}
}

@article{chu2021twins,
  title={Twins: Revisiting the design of spatial attention in vision transformers},
  author={Chu, Xiangxiang and Tian, Zhi and Wang, Yuqing and Zhang, Bo and Ren, Haibing and Wei, Xiaolin and Xia, Huaxia and Shen, Chunhua},
  journal={Advances in neural information processing systems},
  volume={34},
  pages={9355--9366},
  year={2021}
}

@article{chu2021conditional,
  title={Conditional positional encodings for vision transformers},
  author={Chu, Xiangxiang and Tian, Zhi and Zhang, Bo and Wang, Xinlong and Shen, Chunhua},
  journal={arXiv preprint arXiv:2102.10882},
  year={2021}
}

@article{ma2025follow,
  title={Follow-your-motion: Video motion transfer via efficient spatial-temporal decoupled finetuning},
  author={Ma, Yue and Liu, Yulong and Zhu, Qiyuan and Yang, Ayden and Feng, Kunyu and Zhang, Xinhua and Li, Zhifeng and Han, Sirui and Qi, Chenyang and Chen, Qifeng},
  journal={arXiv preprint arXiv:2506.05207},
  year={2025}
}

@article{huang2025taming,
  title={Taming Hallucinations: Boosting MLLMs' Video Understanding via Counterfactual Video Generation},
  author={Huang, Zhe and Wen, Hao and Hao, Aiming and Song, Bingze and Wu, Meiqi and Wu, Jiahong and Chu, Xiangxiang and Lu, Sheng and Wang, Haoqian},
  journal={arXiv preprint arXiv:2512.24271},
  year={2025}
}
